\documentclass[preprint]{article}

\usepackage{microtype}
\usepackage{graphicx}
\usepackage{subfigure}
\usepackage{booktabs} 
\usepackage{soul} 
\usepackage[table]{xcolor} 
\usepackage{multirow} 
\usepackage{tabularx} 

\usepackage{hyperref}

\usepackage{icml2025}

\usepackage{amsmath}
\usepackage{amssymb}
\usepackage{mathtools}
\usepackage{amsthm}

\usepackage[capitalize,noabbrev]{cleveref}

\theoremstyle{plain}
\newtheorem{theorem}{Theorem}[section]

\theoremstyle{definition}

\theoremstyle{remark}

\usepackage{enumitem}
\usepackage[dvipsnames]{xcolor}
\usepackage[most]{tcolorbox}
\usepackage{listings}
\newtcolorbox{dialogbox}[1][]{
  arc=4mm,
  colback=lightgray!20,
  colframe=orange!60!black,
  rounded corners,
  boxrule=1.5pt,
  fonttitle=\sffamily\bfseries,
  coltitle=white,
  toptitle=1mm,
  bottomtitle=1mm,
  title=#1, 
  frame style={dashed}, 
  breakable,  
}
\usepackage[textsize=tiny]{todonotes}

\icmltitlerunning{Surpassing Discrete-Token LLM Reinforcement Learning via Gumbel-Reparameterized Soft-Thinking Policy Optimization}

\begin{document}

\twocolumn[
\icmltitle{SofT-GRPO: Surpassing Discrete-Token LLM Reinforcement Learning via Gumbel-Reparameterized Soft-Thinking Policy Optimization}




\begin{icmlauthorlist}
\icmlauthor{Zhi Zheng}{nus}
\icmlauthor{Yu Gu}{nanjing}
\icmlauthor{Wei Liu}{nus}
\icmlauthor{Yee Whye Teh}{oxford}
\icmlauthor{Wee Sun Lee}{nus}
\end{icmlauthorlist}

\icmlaffiliation{nus}{School of Computing, National University of Singapore, Singapore}
\icmlaffiliation{oxford}{Department of Statistics, University of Oxford, United Kingdom}
\icmlaffiliation{nanjing}{School of Intelligence Science and Technology, Nanjing University, China}

\icmlcorrespondingauthor{Wei Liu}{weiliu87@nus.edu.sg}

\icmlkeywords{Machine Learning, ICML}

\vskip 0.3in
]



\printAffiliationsAndNotice{}  

\begin{abstract}
The soft-thinking reasoning paradigm has demonstrated superior performance over traditional discrete-token Chain-of-Thought (CoT) reasoning in various scenarios. However, while discrete-token CoT reasoning can be reinforced through advanced reinforcement learning with verifiable rewards (RLVR) techniques such as group relative policy optimization (GRPO), extending the soft-thinking reasoning with such strong techniques remains challenging. This difficulty stems from the complexities of injecting stochasticity into soft-thinking tokens and updating soft-thinking policies accordingly. As a result, previous attempts to combine soft-thinking with RLVR typically underperform their discrete-token RLVR counterparts. To fully unlock the potential of soft-thinking, this paper presents a powerful policy optimization algorithm, SofT-GRPO. It injects the Gumbel noise into token probabilities with Gumbel-Softmax for controllable stochasticity, and leverages the Gumbel reparameterization trick to achieve accurate credit assignment to LLM soft-thinking policies. We conduct experiments over LLMs ranging from 1.5B to 7B parameters, where SofT-GRPO enables LLMs with soft-thinking to slightly outperform discrete-token CoT GRPO on Pass@1 (+0.13\% on average accuracy), and brings a substantial uplift on Pass@32 (+2.19\% on average)\footnote[2]{The codes are available at \url{https://github.com/zz1358m/SofT-GRPO-master}}.
\end{abstract}

\begin{figure}[htbp]
    \centering
    \subfigure[LLM Reasoning with Discrete-Token CoT]{\includegraphics[width = 0.48\textwidth]{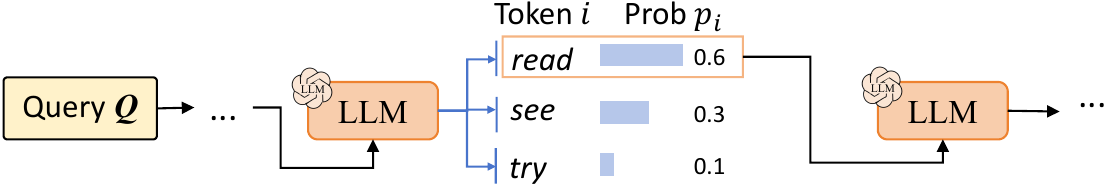}}
    \subfigure[LLM Reasoning with Soft-Thinking \citep{zhang2025soft}]{\includegraphics[width = 0.48\textwidth]{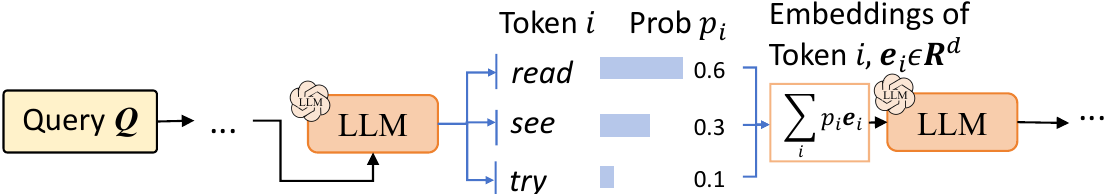}}
    \subfigure[GRPO \citep{shao2024deepseekmath} for Discrete-Token CoT]{\includegraphics[width = 0.48\textwidth]{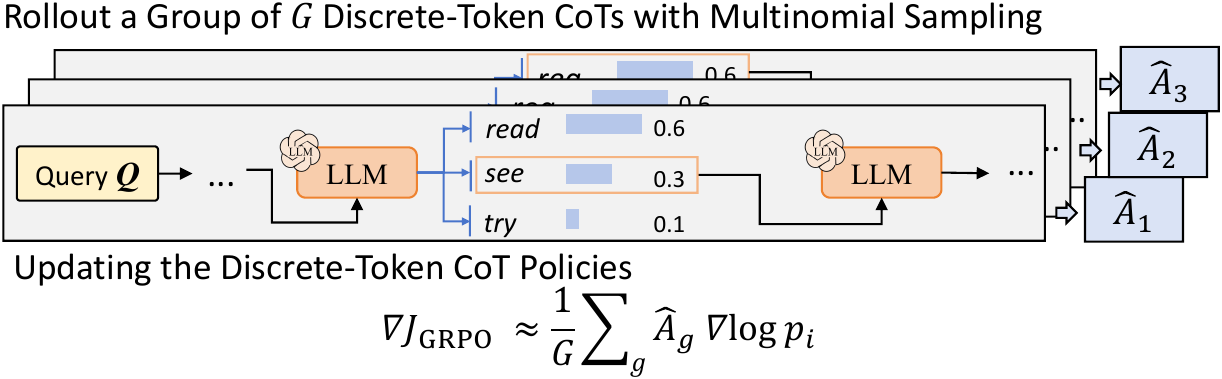}}
    \subfigure[Existing Work of Soft-Thinking + GRPO \citep{butt2025soft}]{\includegraphics[width = 0.48\textwidth]{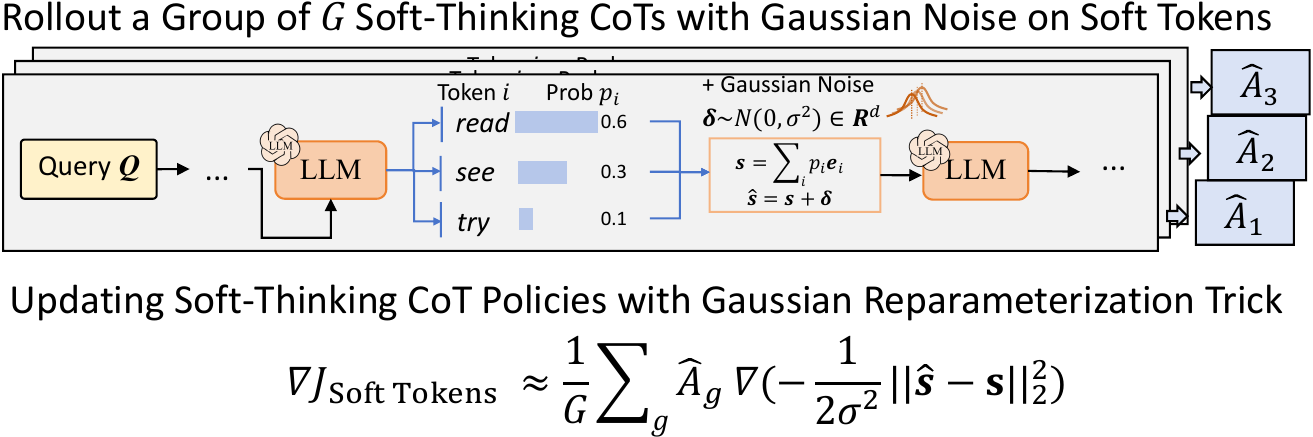}}
    \subfigure[SofT-GRPO (Ours) for Reinforcing Soft-Thinking]{\includegraphics[width = 0.48\textwidth]{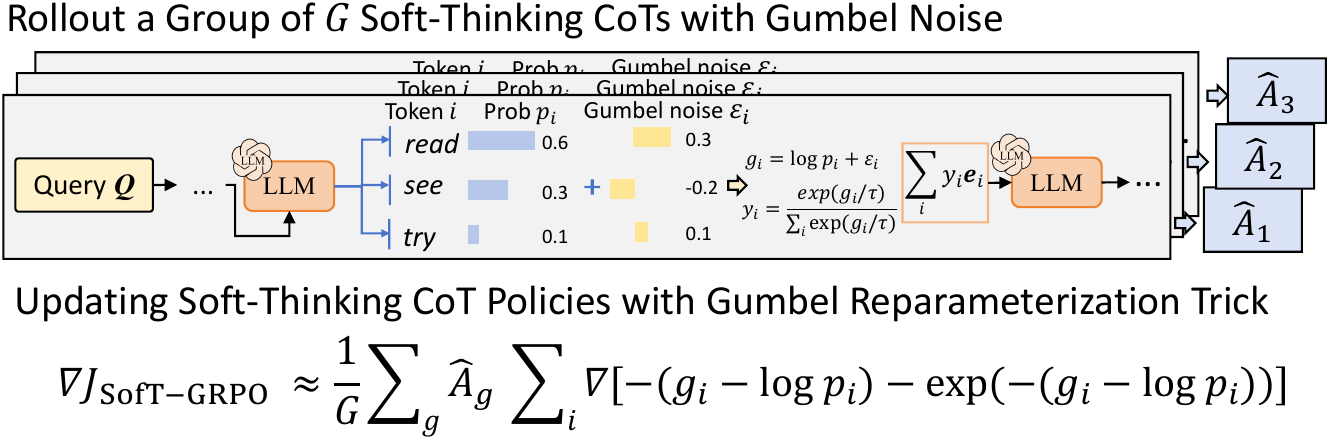}}
    \caption{The soft-thinking pattern (b) passes the expectation of token embeddings to the next LLM step \citep{zhang2025soft}, which can surpass the conventional discrete-token CoT (a) without any fine-tuning. However, employing the GRPO algorithm (c) will boost the performance of discrete-token CoT over soft-thinking, and existing attempts (d) of applying RLVR to soft-thinking derive inferior performances. The proposed SofT-GRPO (e) provides the first powerful RLVR algorithm, which can outperform the discrete-token CoT with GRPO on Pass@1, especially Pass@K. }\label{fig:figure1}
\end{figure}

\vspace{-25pt}
\section{Introduction}
Reasoning with large language models (LLMs) has demonstrated impressive versatility in diverse domains \citep{sprague2024cot}. However, most existing reasoning methods rely on the generation of discrete language tokens, which will limit their ability to represent certain abstract concepts that cannot be accurately represented by a discrete token \citep{zhang2025soft}. In pursuit of better expression of abstract ideas, \citet{zhang2025soft} presents the soft-thinking reasoning pattern. As shown in Figure~\ref{fig:figure1} (b), it replaces each discrete token in the chain-of-thought (CoT) with a continuous representation: the weighted sum of $d$-dimensional token embeddings, taking the output probabilities as the weights. With suitable sampling techniques—such as with the Gumbel-Softmax technique \citep{wu2025llms} or the Dirichlet resampling technique \citep{zhuang2025text}—soft-thinking can outperform discrete-token CoT on a wide range of tasks without requiring any fine-tuning \cite{wu2025llms}.

Recently, reinforcement learning with verifiable rewards (RLVR) approaches have become standard for improving discrete-token CoT reasoning \citep{wang2024reinforcement, liu2025understanding, yu2025dapo}. The most representative RLVR method, Group Relative Policy Optimization (GRPO; Figure~\ref{fig:figure1} (c)) \citep{shao2024deepseekmath}, samples groups of CoT trajectories per query and reinforces the policy toward higher-reward examples. Notably, RLVR fine-tuning can yield discrete-token CoT performance that surpasses that of soft-thinking reasoning from the original, unfine-tuned LLMs. Moreover, after discrete-token GRPO, the model's soft-thinking ability often does not improve as substantially as its discrete-token reasoning ability. 

However, extending RLVR methods to soft-thinking reasoning remains challenging \citep{jain2025learning}. As illustrated in Figure~\ref{fig:figure1} (b), in contrast to the stochastic nature of sampling-based discrete-token CoTs, soft-thinking reasoning produces \textbf{deterministic} soft-tokens, lacking both randomness and gradients for RLVR. So, extending RLVR to soft-thinking should (\textbf{1}) introduce controllable stochasticity to adequately explore diverse soft-thinking paths, and (\textbf{2}) enable unbiased policy updates that properly assign the credits of high-reward soft-thinking samples to LLM logits. Previous work \citep{butt2025soft} (see Figure~\ref{fig:figure1} (d)) attempts to address stochasticity by injecting Gaussian noise into soft tokens and applying a Gaussian reparameterization trick for RLVR gradients, but such approaches have struggled to match the performance of discrete-token RLVR, largely due to \textbf{difficulties in attributing RLVR advantages to logits}.

To address these challenges and fully unlock the LLM reasoning ability by soft-thinking, we propose a powerful policy optimization algorithm, \textbf{SofT-GRPO}. As shown in Figure \ref{fig:figure1}(e), to introduce \textbf{controllable stochasticity} in the rollout process, SofT-GRPO samples groups of soft-thinking reasoning paths by injecting Gumbel noises into the output token probabilities and employs the Gumbel-Softmax technique to avoid invalid soft-tokens outside the pre-trained discrete-token embedding space. To \textbf{accurately assign the reward credits} to the \textbf{output probability} of LLMs in soft-thinking policy updates, SofT-GRPO leverages the reparameterization trick on the Gumbel distribution. We perform thorough evaluations of SofT-GRPO on three representative LLMs from 1.5B to 7B parameters and across five numerical reasoning benchmarks. Experimental results show that SofT-GRPO-enhanced soft-thinking not only surpasses discrete-token GRPO on the averaged Pass@1 but also yields substantial gains on Pass@32, highlighting the practical advantages of robust policy optimization for soft-thinking reasoning. Our contributions are as follows: 
\begin{itemize}[leftmargin=*] 

\item We present \textbf{SofT-GRPO}, a powerful RLVR algorithm designed with the soft-thinking reasoning paradigm. It adopts the Gumbel-Softmax technique into the group rollout process, actively obtaining \textbf{diverse but valid} soft-thinking reasoning paths.

\item SofT-GRPO proposes a novel gradient estimation approach via \textbf{Gumbel reparameterization}, first achieving \textbf{precise attribution} of improvements to the LLM’s output probability distributions in policy optimization.

\item Over eight in-domain and out-of-domain benchmarks, SofT-GRPO improves the LLM's reasoning ability with soft-thinking, \textbf{outperforming the discrete-token GRPO}, especially at higher sample rates (Pass@16 or Pass@32). 

\end{itemize}

\section{Preliminaries}

\subsection{Discrete-Token CoT Reasoning} \label{languageCoT}

To solve a $|\boldsymbol{Q}|$-token question $\boldsymbol{Q}=(q_1,\ldots,q_{|\boldsymbol{Q}|})$, the discrete-token CoT reasoning process generates $|\boldsymbol{R}|$ reasoning CoT tokens $\boldsymbol{R}=(r_1,\ldots,r_{|\boldsymbol{R}|})$ before predicting the answer $\boldsymbol{A}=(a_1,\ldots,a_{|\boldsymbol{A}|})$ \citep{guo2025deepseek,sprague2024cot}. Each of $\boldsymbol{Q}$, $\boldsymbol{R}$, and $\boldsymbol{A}$ is a token sequence over the vocabulary $\mathcal{T}$ (i.e., $\boldsymbol{Q}, \boldsymbol{R}, \boldsymbol{A} \in \mathcal{T}^*$), where $\mathcal{T}$ is the set of all possible language tokens. Language reasoning tokens and answer tokens are generated with the next-token prediction (NTP) policy of LLM $\pi_{\theta}$ as follows:
\begin{equation}
\begin{aligned}
p(\boldsymbol{R},\boldsymbol{A}|\boldsymbol{Q}) = &\prod_{t=1}^{|\boldsymbol{R}|}\pi_{\theta}(r_t|[\boldsymbol{Q},(r_1,\ldots,r_{t-1})])\\& \quad \prod_{t=1}^{|\boldsymbol{A}|}\pi_{\theta}(a_t|[\boldsymbol{Q},\boldsymbol{R},(a_1,\ldots,a_{t-1})]),
\end{aligned}\label{problanguage}
\end{equation}
where $[\cdot,\ \cdot]$ and $[\cdot,\ \cdot,\ \cdot]$ denotes concatenation.

\begin{figure*}[htbp]\vspace{-5pt}
\begin{equation}
\begin{aligned}
&\mathcal{J}_{\text{GRPO}}(\theta) = \frac{1}{G} \mathbb{E}_{\boldsymbol{Q}\sim \mathcal{D}, \{\boldsymbol{R}\}_{g=1}^G, \{\boldsymbol{A}\}_{g=1}^G \sim p(\cdot,\cdot|\boldsymbol{Q})}\\
&\Bigg[\sum_{g=1}^G \frac{1}{\left| \boldsymbol{R}_g \right|+\left| \boldsymbol{A}_g \right|} \sum_{t=1}^{\left| \boldsymbol{R}_g \right|+\left| \boldsymbol{A}_g \right|} \Big( \min \left( p_{g,t} \hat{A}_{g}, \text{clip}(p_{g,t}, 1 - \epsilon, 1 + \epsilon)\hat{A}_{g} \right)  - \beta \mathbb{D}_{KL}(\pi_{\theta} || \pi_{\theta_\text{ref}}) \Bigg]\\
&\qquad\hat{A}_{g}=\frac{f(\boldsymbol{A}_{g})-\text{mean}(f(\boldsymbol{A}))_{g=1}^G}{\text{std}(f(\boldsymbol{A}))_{g=1}^G},\qquad p_{g,t} = 
\begin{cases} 
\frac{\pi_{\theta}(a_{g,t} | [\boldsymbol{Q}, \boldsymbol{R}_g, (a_{g,1}, \ldots, a_{g,t-1})])}{\pi_{\theta_{\text{old}}}(a_{g,t} | [\boldsymbol{Q}, \boldsymbol{R}, (a_{g,1}, \ldots, a_{g,t-1})])} & \text{if } t > |\boldsymbol{R}_g| \\ 
\frac{\pi_{\theta}(r_{g,t} | [\boldsymbol{Q}, (r_{g,1}, \ldots, r_{g,t-1})])}{\pi_{\theta_{\text{old}}}(r_{g,t} | [\boldsymbol{Q}, (r_{g,1}, \ldots, r_{g,t-1})])} & \text{if } t \leq |\boldsymbol{R}_g|.
\end{cases}
\end{aligned}\label{grpo}
\end{equation}\vspace{-5pt}
\end{figure*}

\textbf{Supervised Fine-tuning (SFT) for Discrete-Token Reasoning.} As straightforward fine-tuning methods for LLM reasoning, the SFT methods \citep{wu2025llm, zheng2025reasoning} collect high-quality CoT labels for each question and then fine-tune LLMs for correct predictions. However, SFT methods highly rely on the quality of CoT labels, which are hard to obtain for complex datasets. Moreover, there are also concerns about the out-of-domain generalization ability of SFT \citep{chu2025sft}.

\textbf{RLVR Fine-tuning for Discrete-Token Reasoning.} RLVR fine-tuning methods such as GRPO \citep{liu2024deepseek}, Dr. GRPO \citep{liu2025understanding}, and DAPO \cite{yu2025dapo} sample several CoTs $[\boldsymbol{R}, \boldsymbol{A}]$ and assign a reward to each of them based on the quality of the answers $\boldsymbol{A}$. The standard discrete-token GRPO algorithm \citep{shao2024deepseekmath} samples $G$ CoTs for each question $\boldsymbol{Q}$ with sampling policy $\pi_{\theta_\text{old}}$ and optimizes the reasoning policy towards CoTs with higher rewards. Its loss function is shown in Eq. \eqref{grpo}, where $\hat{A}_{g}$ represents the advantage function for the $g$-th CoT, the reward function $f(\boldsymbol{A}_{g})=1$ if and only if the answer $\boldsymbol{A}_{g}$ is correct, $\epsilon$ is the clipping hyperparameter, $p_{g,t}$ is calculated in an off-policy way for both the current policy $\pi_{\theta}$ and the sampling policy $\pi_{\theta_\text{old}}$, and $\mathbb{D}_{KL}$ represents the KL-divergence of $\pi_{\theta}$ and a reference policy $\pi_{\theta_\text{ref}}$. RLVR methods can improve the discrete-token reasoning performance of LLMs across a wide range of reasoning problems without the need for CoT labels. However, for each CoT in each LLM step, discrete-token GRPO updates the probability of only one token based on the reward. As mentioned in \citet{yue2025does}, making RLVR does not really incentivize reasoning capacity in LLMs beyond the base model.

\subsection{Soft-Thinking Reasoning Paradigm} \label{priliminary-2}
At each of the CoT reasoning steps, conventional discrete-token reasoning is constrained to selecting one token from the token set $\mathcal{T}$, which may hinder the model's ability to express certain abstract concepts that cannot be easily represented by a single deterministic token \citep{zhang2025soft}, thus undermining the LLM's expression ability in reasoning. To represent abstract concepts, \citet{zhang2025soft} presents the soft-thinking paradigm, which replaces discrete-token reasoning steps $\boldsymbol{R}$ with a soft-thinking reasoning path $\boldsymbol{S}=(\boldsymbol{s}_1,\ldots,\boldsymbol{s}_{|\boldsymbol{S}|})$. Each token $\boldsymbol{s}_i\in \mathcal{R}^d$ (noted soft token) is a real vector, which is the sum of token embeddings weighted by their respective output probabilities. They are then fed into the next LLM step as follows:
\begin{equation}
\begin{aligned}
&\boldsymbol{p}_t = \pi_{\theta}(\cdot|[\boldsymbol{Q}, (\boldsymbol{s}_1, \ldots, \boldsymbol{s}_{t-1})])),\\
&\boldsymbol{s}_t = \sum_{i=1}^{|\mathcal{T}|}p_i \cdot \boldsymbol{e}_i,
\end{aligned}\label{raw-soft-thinking}
\end{equation}
where $p_i \in [0,1]$ is the predicted probability of token $i$ in $\boldsymbol{p}_t$ and $\boldsymbol{e}_i \in \mathcal{R}^d$ is the LLM embeddings of token $i$. As detailed in Appendix \ref{Appendixa3}, the soft tokens lie in the \textbf{convex hull} of token embeddings, making it possible to maintain validity while transmitting more information.

Based on Eq. \eqref{raw-soft-thinking}, \citet{wu2025llms} find that most pre-trained LLMs tend to be single-threaded and the vanilla soft-thinking reasoning may lead to \textbf{greedy feedback loop} that suppresses alternative reasoning paths and undermines the benefits of transmitting more informative soft-tokens. So, they propose to introduce sampling strategies in the soft-thinking reasoning process for randomness, employing Gumbel noises and Gumbel-Softmax technique \citep{maddison2016concrete,jang2016categorical} based on the output probability $\boldsymbol{p}_t$ as follows:
\begin{equation}
\begin{aligned}
& g_i = \log p_i + \epsilon_i,
&y_i = \dfrac{\exp (g_i/\tau _g)}{\sum_{i=1}^{|\mathcal{T}|}\exp (g_i/\tau_g)},\\ 
&\boldsymbol{s}_t = \sum_{i=1}^{|\mathcal{T}|}y_i \cdot \boldsymbol{e}_i,
\end{aligned}\label{gumbel-soft-thinking}
\end{equation}
where $\epsilon_i$ is a scalar noise sampled from the Gumbel distribution Gumbel$(0,1)$, and $\tau_g$ is the temperature of Gumbel-Softmax. Besides Gumbel-Softmax, \citet{wu2025llms} also tries to use the Dirichlet resampling technique as follows:
\begin{equation}
\begin{aligned}
 (x_1,\ldots,x_{|\mathcal{T}|})\sim \text{Dirichlet}(\alpha, \boldsymbol{p}_t),\quad \boldsymbol{s}_t = \sum_{i=1}^{|\mathcal{T}|}x_i \cdot \boldsymbol{e}_i,
\end{aligned}\label{dirichlet-soft-thinking}
\end{equation}
where $\alpha$ is a scaling parameter. Empirically, incorporated with Gumbel noises and the Gumbel-Softmax technique, the soft-thinking pattern can outperform conventional discrete-token CoT on a broad range of tasks, including numerical, code, and scientific reasoning, without requiring any fine-tuning. \citep{wu2025llms}.

\begin{figure*}[htbp]
\centering\vspace{-2pt}
\includegraphics[width=\linewidth]{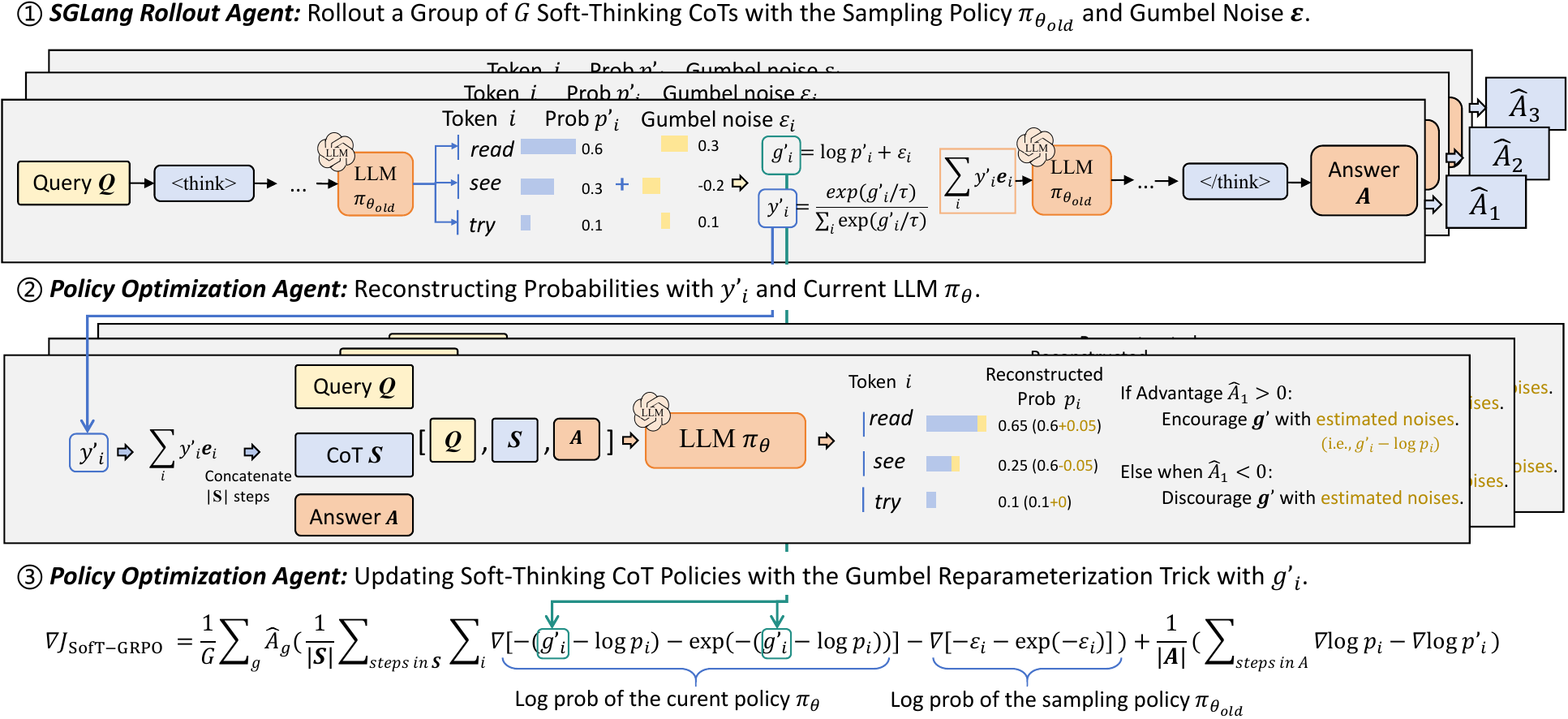}
\caption{The pipeline of the proposed SofT-GRPO algorithm. In training with each Query $\boldsymbol{Q}$, the SofT-GRPO first generates a group of $G$ soft-thinking reasoning paths with Gumbel noises and the Gumbel-Softmax technique \cite{wu2025llms}. We transmit the value $g'_i$ and $y'_i$ for the loss calculation afterward. Then, we reconstruct the soft-thinking input. Finally, we update the soft-thinking policy with the off-policy proximal policy optimization \citep{schulman2017proximal}, optimizing soft-thinking policies with Gumbel reparameterization.}
\label{fig:main}\vspace{-2pt}
\end{figure*}

\subsection{Related Work on RLVR Soft-Thinking}

However, after being boosted with the RLVR fine-tuning, discrete-token reasoning will clearly outperform soft-thinking. So, \citet{butt2025soft} tries to similarly improve the soft-thinking reasoning ability using GRPO by adding Gaussian noise on the soft-token $\boldsymbol{s}_t$ in Eq. \eqref{raw-soft-thinking} as follows:
\begin{equation}
    \hat{\boldsymbol{s}}_t =\boldsymbol{s}_t+\mathcal{N}(0,\sigma^2 Id).
\end{equation}
In deriving the soft-thinking probability in GRPO, they restore the value of $\hat{\boldsymbol{s}}_t$ in the rollout and calculate the log probability for the soft-thinking reasoning path $\boldsymbol{S}$ with the Gaussian reparameterization trick as follows, keeping the updating process of the answering part $\boldsymbol{A}$ unchanged:
\begin{equation}
p(\hat{\boldsymbol{s}}_t) \propto \exp\left( -\frac{1}{2\sigma^2}||\hat{\boldsymbol{s}}_t-\boldsymbol{s}_t||^2_2\right).
\end{equation}
Although this method can improve the soft-thinking ability of base LLMs and show improvements compared to discrete-token GRPO on the Pass@32 metrics (i.e., the pass rate with 32 attempts), there is a severe degradation in the average accuracy. This is mainly due to the following two drawbacks: \textbf{(1)} Adding noises on soft-tokens may help avoid inputs outside the pre-trained discrete-token embedding space, but adding noise to soft-tokens instead of logits is not direct and may theoretically mismatch the LLM predictions. As analyzed in Appendix \ref{Appendixc1}, token embeddings are linearly dependent ($|\mathcal{T}|\gg d$), so they \textbf{cannot attribute which probability $p_i$ is beneficial} to the effectiveness of added noise. Moreover, the added noise may even be impossible to be represented by the embeddings. \textbf{(2)} They do \textbf{not use advanced sampling methods} for effectiveness in training (e.g., incorporating Dirichlet resampling or Gumbel-Softmax technique \citep{wu2025llms}), which will undermine the performance of sampled soft-thinking reasoning paths. 

As a result, existing attempts of applying GRPO to soft-thinking will not keep its advantage over the discrete-token CoT under the no-finetune setting.

\begin{figure*}[htbp]
\begin{equation}
\begin{aligned}
&\mathcal{J}_{\text{SofT-GRPO}}(\theta) = \frac{1}{G} \mathbb{E}_{\boldsymbol{Q}\sim \mathcal{D}, \{\boldsymbol{S}\}_{g=1}^G, \{\boldsymbol{A}\}_{g=1}^G \sim p(\cdot,\cdot|\boldsymbol{Q})}\\
&\Bigg[\sum_{g=1}^G \frac{1}{\left| \boldsymbol{S}_g \right|+\left| \boldsymbol{A}_g \right|} \sum_{t=1}^{\left| \boldsymbol{S}_g \right|+\left| \boldsymbol{A}_g \right|} \Big( \min \left( p_{g,t} \hat{A}_{g}, \text{clip}(p_{g,t}, 1 - \epsilon, 1 + \epsilon)\hat{A}_{g} \right)  - \beta \mathbb{D}_{KL}(\pi_{\theta} || \pi_{\theta_\text{ref}}) \Bigg]\\
& \ p_{g,t} = 
\begin{cases} 
\frac{\pi_{\theta}(a_{g,t} | [\boldsymbol{Q}, \boldsymbol{S}, (a_{g,1}, \ldots, a_{g,t-1})])}{\pi_{\theta_{\text{old}}}(a_{g,t} | [\boldsymbol{Q}, \boldsymbol{S}, (a_{g,1}, \ldots, a_{g,t-1})])} & \text{if } t > |\boldsymbol{S}_g| \\ 
\exp\left(\sum_{i=1}^{|\mathcal{T}|}\big(-(g'_i-\log p_i)-\exp(-g'_i+\log p_i)\big)-\big(-\epsilon_i-\exp(-\epsilon_i)\big)\right) & \text{if } t \leq |\boldsymbol{S}_g|.
\end{cases}
\end{aligned}\label{soft-grpo}
\end{equation}\vspace{-10pt}
\end{figure*}

\section{SofT-GRPO: Reinforcing Soft-Thinking Policy with Gumbel Reparameterization}

To fully unlock the potential of soft-thinking for better LLM reasoning, we present the RLVR algorithm SofT-GRPO. As shown in Figure \ref{fig:main}, it first samples soft-thinking reasoning paths with controllable randomness using the Gumbel-Softmax technique \citep{wu2025llms}. In the following policy update stage, we propose a novel SofT-GRPO loss function with the Gumbel reparameterization trick, which accurately assigns reward credits to token probabilities.

\subsection{Group Rollout with Gumbel Noise}

As an off-policy RLVR algorithm, SofT-GRPO samples a group of $G$ soft-thinking CoTs for each query $\boldsymbol{Q}$ during rollout and saves them in a replay buffer. A principal challenge for applying RLVR to soft-thinking is that vanilla soft-thinking generates deterministic soft tokens, as each $\boldsymbol{s}_t$ is a fixed weighted sum of embeddings based solely on model probabilities (Eq.~\eqref{raw-soft-thinking}). So, to explore \textbf{diverse and powerful} soft-thinking reasoning paths in the rollout process, we introduce the Gumbel-Softmax resampling technique (shown in Eq. \eqref{gumbel-soft-thinking}) as follows:
\begin{equation}
    \begin{aligned}
    &(p_1,\ldots,p_{|\mathcal{T}|}) = \pi_{\theta_\text{old}}(\cdot| [\boldsymbol{Q}, (\boldsymbol{s}_{1}, \ldots, \boldsymbol{s}_{t-1})]),\\
    &g'_i=\log\ p_i+\epsilon_i,\quad  y'_i = \dfrac{\exp(g'_i/\tau _g)}{\sum_{i=1}^{|\mathcal{T}|}\exp(g'_i/\tau _g)},\\
    &\boldsymbol{s}_t = \sum_{i=1}^{|\mathcal{T}|}y'_i\cdot \boldsymbol{e}_i,
    \end{aligned}\label{soft-grporollout}
\end{equation}

where $\boldsymbol{e}_i$ is the embedding of the $i$-th token, $\pi_{\theta_\text{old}}$ is the sampling policy used in rollout, and $\epsilon_i$ is a scalar Gumbel noise. Using the inverse transform sampling \citep{jang2016categorical}, we sample $\epsilon_i=-\log(-\log(u))$ where $u\sim \text{Uniform}(0, 1)$. Besides bringing stochasticity, as discussed in Section \ref{priliminary-2}, resampling with Gumbel-Softmax will bring potential improvements on the quality of reasoning paths. 

Due to limitations in the amount of fine-tuning data, fine-tuning LLMs with soft-thinking needs to preserve the pre-trained knowledge. So, applying RLVR to soft-thinking should also keep the generated soft tokens $\boldsymbol{s}_t$ \textbf{within the pre-trained discrete-token embedding space}. In contrast to noise injection based on other distributions (e.g., Dirichlet or Gaussian), Gumbel-Softmax can better maintain the stability of the soft-token embeddings when providing stochasticity. As established in Theorem~\ref{theory31} (see Appendix~\ref{Appendixc2} for proof), it produces a relaxation that maintains samples close to the discrete token distribution, thus mitigating the risk of generating out-of-vocabulary soft tokens.
\begin{theorem}[Gumbel-max Trick]
Let $(p_1, \dots, p_n)$ be nonnegative, and $\epsilon_1, \dots, \epsilon_n$ independent samples from $\mathrm{Gumbel}(0,1)$ \citep{maddison2016concrete},
\begin{equation}
    \Pr\left(
        j = \arg\max_{i}\left(\log p_i + \epsilon_i\right)
    \right) =
    \frac{p_j}{\sum_{i=1}^n p_i}.
\end{equation}\label{theory31}\vspace{-10pt}
\end{theorem}

\subsection{Gumbel Reparameterization for Loss Function}\label{Gumbelre}

In RLVR with discrete tokens (Eq. \eqref{problanguage}), the probability of CoT trajectories can be obtained from the output Categorical distributions. Calculating such a probability is impossible in vanilla soft-thinking. To address this, SofT-GRPO proposes using the Gumbel reparameterization trick after obtaining the Gumbel noise $\boldsymbol{\epsilon}$ during the rollout process. We take $\boldsymbol{\epsilon}$ as RLVR actions, consider transforming $\boldsymbol{\epsilon}$ to soft tokens $\boldsymbol{s}_t$ as a part of the transition, and estimate the next-soft-token probability with the density of Gumbel noise as follows:
\begin{equation}
    p(\boldsymbol{g}'| [\boldsymbol{Q}, (\boldsymbol{s}_{1}, \ldots, \boldsymbol{s}_{t-1})], \theta_\text{old})=\exp\big[\sum_{i=1}^{|\mathcal{T}|}-\epsilon_i-\exp(-\epsilon_i)\big],\label{repa}
\end{equation}
where $\boldsymbol{g}'=(g'_1,\ldots,g'_{|\mathcal{T}|})$ is the $g'_i$ values in Eq. \eqref{soft-grporollout} collected from the replay buffer. We provide the derivation for this equation in Appendix \ref{Appendixc3}. Following GRPO, SofT-GRPO requires off-policy probabilities for clipping. When calculating off-policy probabilities, we first reconstruct the soft-thinking reasoning paths $\boldsymbol{S}=(\boldsymbol{s}_1,\ldots,\boldsymbol{s}_{|\boldsymbol{S}|})$ from the restored $y'_i$ values in the rollout process. Then, the probability is calculated with the restored $g'_i$ values as follows:
\begin{equation}
\begin{aligned}
    &p(\boldsymbol{g}'| [\boldsymbol{Q}, (\boldsymbol{s}_{1}, \ldots, \boldsymbol{s}_{t-1})],\theta)=\\
    &\qquad \exp\left(\sum_{i=1}^{|\mathcal{T}|}-(g'_i-\log p_i)-\exp(-(g'_i-\log p_i))\right),\\
    &\text{where} \  (p_1,\ldots,p_{|\mathcal{T}|}) = \pi_{\theta}(\cdot| [\boldsymbol{Q}, (\boldsymbol{s}_{1}, \ldots, \boldsymbol{s}_{t-1})]).\label{repac}
\end{aligned}
\end{equation}
Finally, the SofT-GRPO loss function is Eq. \eqref{soft-grpo}, where $\pi$ and $\pi_\text{ref}$ are token probability distributions that take history soft-tokens $\boldsymbol{S}$ and answer tokens $\boldsymbol{A}$ as input. $p_{g,t}$ is calculated as the probability ratio between current policy $p_\theta$ (Eq. \eqref{repa}) and sampling policy $p_{\theta_\text{old}}$ (Eq. \eqref{repac}). Intuitively, SofT-GRPO recalculates the noise corresponding to $\boldsymbol{g}'$ obtained in rollout and encourages this noise only when the current soft-thinking trajectory leads to a higher $\hat{A}_g$. Unlike discrete-token GRPO in Eq. \eqref{grpo}, SofT-GRPO updates the probability of several tokens at each LLM step $t$.

\begin{table*}[t]
\centering
\renewcommand\arraystretch{1.05}
\setlength{\tabcolsep}{2.2mm}
\caption{Experiment results of baselines and the proposed SofT-GRPO on five numerical reasoning benchmarks. We cover 3 LLMs from 1.5B to 7B and two reasoning patterns, i.e., discrete-token CoT and soft-thinking reasoning. @1 metrics denote the Mean@32 values, where we run each method 32 times on the dataset for average Pass@1 accuracies. @16 and @32 denote the Pass@16 and Pass@32 values on the dataset, respectively. We multiply all the results by 100 to highlight the differences between results. The best result on each metric and dataset is \underline{underlined}, the best average result is \textbf{bolded}, and the second-best average result is shaded.}
\resizebox{\textwidth}{!}{\vspace{-10pt}
\begin{tabular}{lcccccccccccccccccc}
\bottomrule[0.5mm]
\multicolumn{1}{c|}{Dataset}& \multicolumn{3}{c|}{AIME2024} & \multicolumn{3}{c|}{AIME2025} & \multicolumn{3}{c|}{AMC23}    & \multicolumn{3}{c|}{MATH-500} & \multicolumn{3}{c|}{GSM8K}    & \multicolumn{3}{c}{Average}    \\ \hline
\multicolumn{1}{c|}{Metrics}& @1   & @16   & \multicolumn{1}{c|}{@32}    & @1   & @16   & \multicolumn{1}{c|}{@32}    & @1   & @16   & \multicolumn{1}{c|}{@32}    & @1   & @16   & \multicolumn{1}{c|}{@32}    & @1   & @16   & \multicolumn{1}{c|}{@32}    & @1 & @16 & @32  \\ \hline\hline
\multicolumn{19}{c}{\textit{\textbf{DeepSeek-R1-Distill-Qwen-1.5B Base LLM}}}   \\ \hline
\multicolumn{19}{c}{Discrete-Token CoT Reasoning Pattern}  \\ \hline
\multicolumn{1}{l|}{No-Finetune} & 30.6& 70.0& \multicolumn{1}{c|}{73.3}& 23.0& 46.7& \multicolumn{1}{c|}{\underline{53.3}} & 70.7& 92.5& \multicolumn{1}{c|}{95.0}& 84.6& \underline{97.8} & \multicolumn{1}{c|}{97.8}& 81.5& 95.8& \multicolumn{1}{c|}{96.7}& 58.09    & 80.54    & 83.23    \\
\multicolumn{1}{l|}{+ GRPO}& 31.8& 66.7& \multicolumn{1}{c|}{76.7}& 25.3& 46.7& \multicolumn{1}{c|}{46.7}& \underline{77.3} & 95.0& \multicolumn{1}{c|}{95.0}& \underline{87.1} & 97.4& \multicolumn{1}{c|}{97.8}& 84.9& 95.1& \multicolumn{1}{c|}{95.8}& \cellcolor[HTML]{D0CECE}61.28 & 80.16    & 82.39 \\ \hline
\multicolumn{19}{c}{Soft-Thinking Reasoning Pattern}\\ \hline
\multicolumn{1}{l|}{No-Finetune} & 27.3& 66.7& \multicolumn{1}{c|}{70.0}& 23.8& 46.7& \multicolumn{1}{c|}{\underline{53.3}} & 69.9& 95.0& \multicolumn{1}{c|}{95.0}& 79.4& 93.2& \multicolumn{1}{c|}{96.6}& 81.0& 94.6& \multicolumn{1}{c|}{\underline{97.1}} & 56.28    & 79.23    & 82.41    \\
\multicolumn{1}{l|}{+ GRPO}& 29.2& 70.0& \multicolumn{1}{c|}{73.3}& 25.4& 46.7& \multicolumn{1}{c|}{\underline{53.3}} & 75.8& 95.0& \multicolumn{1}{c|}{95.0}& 86.3& 96.8& \multicolumn{1}{c|}{\underline{98.2}} & 84.9& 95.6& \multicolumn{1}{c|}{96.4}& 60.31    & \cellcolor[HTML]{D0CECE}80.81 & \cellcolor[HTML]{D0CECE}83.26    \\
\multicolumn{1}{l|}{+ SofT-GRPO}  & \underline{32.6} & \underline{76.7} & \multicolumn{1}{c|}{\underline{80.0}} & \underline{26.1} & \underline{50.0} & \multicolumn{1}{c|}{\underline{53.3}} & 76.4& \underline{97.5} & \multicolumn{1}{c|}{\underline{97.5}} & 86.3& 97.4& \multicolumn{1}{c|}{98.0}& \underline{85.5} & \underline{96.1} & \multicolumn{1}{c|}{97.0}& \textbf{61.39}  & \textbf{83.54}  & \textbf{85.18}  \\ \hline\hline
\multicolumn{19}{c}{\textit{\textbf{LLaMA-3.2-3B-Instruct Base LLM}}}    \\ \hline
\multicolumn{19}{c}{Discrete-Token CoT Reasoning Pattern}  \\ \hline
\multicolumn{1}{l|}{No-Finetune} & 4.4 & 20.0& \multicolumn{1}{c|}{\underline{26.7}} & 0.3 & 0.3 & \multicolumn{1}{c|}{1.0} & 18.3& 65.0& \multicolumn{1}{c|}{\underline{75.0}} & 38.1& 75.6& \multicolumn{1}{c|}{\underline{84.0}} & 67.9& 92.1& \multicolumn{1}{c|}{94.6}& 25.79    & 50.61    & 56.26    \\
\multicolumn{1}{l|}{+ GRPO}& 7.3 & \underline{23.3} & \multicolumn{1}{c|}{\underline{26.7}} & 0.5 & 3.3 & \multicolumn{1}{c|}{3.3} & 27.3& 62.5& \multicolumn{1}{c|}{67.5}& \underline{48.3} & 77.2& \multicolumn{1}{c|}{82.6}& \underline{79.6} & 95.4& \multicolumn{1}{c|}{96.5}& \cellcolor[HTML]{D0CECE}32.60 & 52.35    & 55.32    \\ \hline
\multicolumn{19}{c}{Soft-Thinking Reasoning Pattern}\\ \hline
\multicolumn{1}{l|}{No-Finetune} & 3.4 & 16.7& \multicolumn{1}{c|}{16.7}& 0.2 & 6.7 & \multicolumn{1}{c|}{6.7} & 17.6& \underline{70.0} & \multicolumn{1}{c|}{77.5}& 36.7& 76.0& \multicolumn{1}{c|}{81.4}& 66.9& 91.6& \multicolumn{1}{c|}{94.7}& 24.96    & 52.18    & 55.39    \\
\multicolumn{1}{l|}{+ GRPO}& \underline{8.0}  & 20.0& \multicolumn{1}{c|}{23.3}& \underline{0.7}  & 3.3 & \multicolumn{1}{c|}{10.0}& 27.3& \underline{70.0} & \multicolumn{1}{c|}{\underline{75.0}} & 47.8& 76.8& \multicolumn{1}{c|}{81.8}& 79.2& 94.8& \multicolumn{1}{c|}{96.3}& \cellcolor[HTML]{D0CECE}32.60 & \cellcolor[HTML]{D0CECE}53.00 & \textbf{57.28}  \\
\multicolumn{1}{l|}{+ SofT-GRPO}  & 7.7 & \underline{23.3} & \multicolumn{1}{c|}{\underline{26.7}} & 0.3 & \underline{10.0} & \multicolumn{1}{c|}{\underline{10.0}} & \underline{31.3} & 67.5& \multicolumn{1}{c|}{67.5}& 47.2& \underline{77.6} & \multicolumn{1}{c|}{83.4}& 77.6& \underline{96.4} & \multicolumn{1}{c|}{\underline{97.7}} & \textbf{32.83}  & \textbf{54.96}  & \cellcolor[HTML]{D0CECE}57.06 \\ \hline\hline
\multicolumn{19}{c}{\textit{\textbf{DeepSeek-R1-Distill-Qwen-7B Base LLM}}}     \\ \hline
\multicolumn{19}{c}{Discrete-Token CoT Reasoning Pattern}  \\ \hline
\multicolumn{1}{l|}{No-Finetune} & \underline{55.7} & \underline{80.0} & \multicolumn{1}{c|}{80.0}& 39.4& \underline{66.7} & \multicolumn{1}{c|}{66.7}& 89.9& \underline{97.5} & \multicolumn{1}{c|}{\underline{97.5}} & \underline{93.6} & \underline{98.8} & \multicolumn{1}{c|}{\underline{99.2}} & 89.5& 96.7 & \multicolumn{1}{c|}{97.0}& 73.62    & \textbf{87.94}  & 88.07    \\
\multicolumn{1}{l|}{+ GRPO}& 54.0& \underline{80.0} & \multicolumn{1}{c|}{80.0}& \underline{40.4} & \underline{66.7} & \multicolumn{1}{c|}{70.0}& 89.2& 95.0& \multicolumn{1}{c|}{95.0}& 93.5& 98.4& \multicolumn{1}{c|}{99.0}& 91.4& 96.1& \multicolumn{1}{c|}{96.6}& 73.69 & 87.24    & 88.12    \\ \hline
\multicolumn{19}{c}{Soft-Thinking Reasoning Pattern}\\ \hline
\multicolumn{1}{l|}{No-Finetune} & 55.3& \underline{80.0} & \multicolumn{1}{c|}{\underline{83.3}} & 39.2& \underline{66.7} & \multicolumn{1}{c|}{70.0}& \underline{90.2} & 95.0& \multicolumn{1}{c|}{\underline{97.5}} & 93.3& \underline{98.8} & \multicolumn{1}{c|}{99.0}& 89.3& 96.6& \multicolumn{1}{c|}{97.0} & 73.44    & \cellcolor[HTML]{D0CECE}87.41 & \cellcolor[HTML]{D0CECE}89.38 \\
\multicolumn{1}{l|}{+ GRPO}& 55.5& \underline{80.0} & \multicolumn{1}{c|}{80.0}& 37.8& 63.3& \multicolumn{1}{c|}{66.7}& \underline{90.1} & \underline{97.5} & \multicolumn{1}{c|}{\underline{97.5}} & \underline{93.6} & 98.6& \multicolumn{1}{c|}{99.0}& 91.8& 96.7& \multicolumn{1}{c|}{96.9}& \textbf{73.76} & 87.23    & 88.01    \\
\multicolumn{1}{l|}{+ SofT-GRPO}  & 53.2& \underline{80.0} & \multicolumn{1}{c|}{\underline{83.3}} & \underline{40.4} & 60.0& \multicolumn{1}{c|}{\underline{73.3}} & 89.6& \underline{97.5} & \multicolumn{1}{c|}{\underline{97.5}} & 93.3& 98.6& \multicolumn{1}{c|}{99.0}& \underline{92.1} & \underline{97.2} & \multicolumn{1}{c|}{\underline{97.7}} & \cellcolor[HTML]{D0CECE}73.74  & 86.66    & \textbf{90.18}  \\ \toprule[0.5mm]
\end{tabular}
}\vspace{-10pt}
\label{maindata}
\end{table*}

\vspace{-2pt}
\section{Experiments}\label{experiment-section}

In this section, we implement the proposed SofT-GRPO algorithm to reinforce the soft-thinking reasoning of three LLMs, including \textit{DeepSeek-R1-Distill-Qwen-1.5B}, \textit{LLaMA-3.2-3B-Instruct}, and \textit{DeepSeek-R1-Distill-Qwen-7B}.

\vspace{-1pt}
\subsection{Implementation Details}

\textbf{Detailed Settings.} In practice, we follow the general setting in soft-thinking works \citep{wu2025llms}, enabling both the top-p and top-k sampling strategies, considering only tokens with $k$ highest probabilities instead of the whole token set $|\mathcal{T}|$ and normalizing their probabilities. In both training and inference, we set top-p as $0.95$ and top-k as $5$, the temperature of LLMs $\tau=1$, and the temperature in Gumbel-Softmax $\tau_g=0.1$.

\textbf{Training \& Testing Settings.} We employ DeepScaler \citep{deepscaler2025} as the training dataset, which contains 40,315 queries. In implementing SofT-GRPO, we use SGLang \citep{zheng2024sglang} for the rollout process and the verl-0.4.x framework \citep{sheng2024hybridflow} for RLVR. 

We involve five numerical reasoning benchmarks as in-domain test sets (AIME2024, AIME2025, AMC23, MATH-500, and GSM8K). We also employ a scientific reasoning benchmark (GPQA Diamond) and two code benchmarks (HumanEval and MBPP) for out-of-domain evaluation. The maximum generation length is 8192 in training and 32768 in testing. Test answers are verified using the Math Verify package \citep{Kydlicek_Math-Verify_Math_Verification}. All experiments are implemented on a node of 8× NVIDIA H200 GPUs (141 GB VRAM each), and detailed parameters are shown in Appendix \ref{Appendixc2}.

\begin{table*}[t]
\centering
\renewcommand\arraystretch{1.05}
\setlength{\tabcolsep}{2.8mm}
\caption{Average accuracies on out-of-domain datasets. We cover GPQA Diamond, HumanEval, and MBPP. @1 metrics denote the Mean@32 values, where we run the methods 32 times on the dataset for average Pass@1 accuracies. @8, @16, and @32 denote the Pass@8, Pass@16, and Pass@32 values on the dataset, respectively.}
\resizebox{\textwidth}{!}{
\begin{tabular}{lcccccccccccccccc}
\bottomrule[0.5mm]
\multicolumn{1}{c|}{Dataset}& \multicolumn{4}{c|}{GPQA Diamond}  & \multicolumn{4}{c|}{HumanEval}     & \multicolumn{4}{c|}{MBPP}    & \multicolumn{4}{c}{Average}   \\ \hline
\multicolumn{1}{c|}{Metrics}& @1   & @8   & @16  & \multicolumn{1}{c|}{@32}  & @1   & @8   & @16  & \multicolumn{1}{c|}{@32}  & @1   & @8   & @16  & \multicolumn{1}{c|}{@32}  & @1    & @8    & @16   & @32   \\ \hline\hline
\multicolumn{17}{c}{\textit{\textbf{DeepSeek-R1-Distill-Qwen-1.5B Base LLM}}}    \\ \hline
\multicolumn{17}{c}{Discrete-Token CoT Reasoning Pattern}\\ \hline
\multicolumn{1}{l|}{No-Finetune} & 36.7 & \underline{84.3} & \underline{92.4} & \multicolumn{1}{c|}{96.0} & 68.1 & 87.2 & 90.2 & \multicolumn{1}{c|}{93.9} & 65.5 & 84.8 & 89.1 & \multicolumn{1}{c|}{91.1} & 56.77 & 85.45 & \cellcolor[HTML]{D0CECE}90.59 & \cellcolor[HTML]{D0CECE}93.64 \\
\multicolumn{1}{l|}{+ GRPO} & 35.4 & 77.8 & 88.4 & \multicolumn{1}{c|}{93.4} & \underline{72.2} & \underline{90.9} & \underline{92.7} & \multicolumn{1}{c|}{94.5} & 68.1 & 85.2 & 87.2 & \multicolumn{1}{c|}{90.0} & 58.56 & 84.62 & 89.41 & 92.65 \\ \hline
\multicolumn{17}{c}{Soft-Thinking Reasoning Pattern}     \\ \hline
\multicolumn{1}{l|}{No-Finetune} & 36.0 & 83.8 & 91.9 & \multicolumn{1}{c|}{\underline{97.0}} & 67.2 & 89.6 & 91.5 & \multicolumn{1}{c|}{92.7} & 64.7 & 84.4 & 86.0 & \multicolumn{1}{c|}{88.7} & 55.98 & \cellcolor[HTML]{D0CECE}85.97 & 89.79 & 92.79 \\
\multicolumn{1}{l|}{+ GRPO} & 36.5 & 81.3 & 91.4 & \multicolumn{1}{c|}{94.4} & 71.8 & 89.0 & \underline{92.7} & \multicolumn{1}{c|}{\underline{95.1}} & 68.1 & 84.8 & 87.9 & \multicolumn{1}{c|}{90.3} & \cellcolor[HTML]{D0CECE}58.82 & 85.05 & \textbf{90.68}    & 93.28 \\
\multicolumn{1}{l|}{+ SofT-GRPO}  & \underline{37.3} & 82.8 & 89.9 & \multicolumn{1}{c|}{95.5} & 71.2 & 88.4 & 91.5 & \multicolumn{1}{c|}{94.5} & \underline{68.8} & \underline{88.7} & \underline{90.3} & \multicolumn{1}{c|}{\underline{91.4}} & \textbf{59.08}    & \textbf{86.65}    & 90.54 & \textbf{93.80}    \\ \toprule[0.5mm]
\end{tabular}
}
\label{ood}
\end{table*}

\begin{table*}[t]
\centering
\renewcommand\arraystretch{1.05}
\setlength{\tabcolsep}{1.2mm}
\caption{Experiments on using majority voting to boost the performances on AIME2024, AIME2025, AMC23 and GSM8K. @1 represents the Pass@1 result from Table \ref{maindata}, which is averaged over 32 runs. M@16 and M@32 represent Major@16 and Major@32, respectively.}
\resizebox{\textwidth}{!}{
\begin{tabular}{l|ccc|ccc|ccc|ccc|ccc|ccc}
\bottomrule[0.5mm]
\multicolumn{1}{c|}{Dataset}      & \multicolumn{3}{c|}{AIME2024}     & \multicolumn{3}{c|}{AIME2025}     & \multicolumn{3}{c|}{AMC23}        & \multicolumn{3}{c|}{MATH-500}        & \multicolumn{3}{c|}{GSM8K} & \multicolumn{3}{c}{Average}       \\\hline
\multicolumn{1}{c|}{Metrics}      & @1         & {\footnotesize M@16}       & \multicolumn{1}{c|}{{\footnotesize M@32}}       & @1         & {\footnotesize M@16}       & \multicolumn{1}{c|}{{\footnotesize M@32}}       & @1         & {\footnotesize M@16}       & \multicolumn{1}{c|}{{\footnotesize M@32}}       & @1         & {\footnotesize M@16}       & \multicolumn{1}{c|}{{\footnotesize M@32}}       & @1         & {\footnotesize M@16}       & {\footnotesize M@32} & @1& {\footnotesize M@16}          & {\footnotesize M@32}          \\
\hline\hline
\multicolumn{19}{c}{\textit{\textbf{DeepSeek-R1-Distill-Qwen-1.5B Base LLM}}}     \\
\hline
\multicolumn{19}{c}{Discrete-Token CoT Reasoning Pattern} \\
\hline
No Fine-tune & 30.6 & 56.7 & 60.0 & 23.0 & \underline{36.7} & \underline{40.0} & 70.7 & 90.0 & \underline{95.0} & 84.6 & 92.6 & \underline{92.4} & 81.5 & 89.5 & 89.5 & 58.1 & 73.1 & 75.4 \\
+ GRPO & 31.8 & 53.3 & 50.0 & 25.3 & 36.7 & 30.0 & \underline{77.3} & 92.5 & 92.5 & \underline{87.1} & 91.6 & 91.4 & 84.9 & 90.1 & 90.4 & 61.3 & 72.8 & 70.9 \\
\hline
\multicolumn{19}{c}{Soft-Thinking Reasoning Pattern}      \\
\hline
No Fine-tune & 27.3 & 56.7 & 60.0 & 23.8 & 33.3 & 33.3 & 69.9 & \underline{95.0} & 92.5 & 79.4 & 88.8 & 89.8 & 81.0 & 89.3 & 89.3 & 56.3 & 72.6 & 73.0 \\
+ GRPO & 29.2 & 43.3 & 46.7 & 25.4 & 30.0 & 30.0 & 75.8 & 92.5 & 92.5 & 86.3 & 91.0 & 91.4 & 84.9 & 90.5 & \underline{90.8} & 60.3 & 69.5 & 70.3 \\
+ SofT-GRPO & \underline{32.6} & \underline{60.0} & \underline{63.3} & \underline{26.1} & 33.3 & 36.7 & 76.4 & 92.5 & \underline{95.0} & 86.3 & \underline{92.2} & 92.0 & \underline{85.5} & \underline{90.6} & 90.5 & \textbf{61.4} & \textbf{73.7} & \textbf{75.5} \\
\toprule[0.5mm]
\end{tabular}
}\vspace{-5pt}
\label{majority voting}
\end{table*}

\textbf{Baselines.} We use both conventional \textbf{discrete-token CoT} reasoning and \textbf{soft-thinking} reasoning as baselines for SofT-GRPO. For discrete-token CoT, we include \textbf{discrete-token GRPO} trained on the same dataset and \textbf{base LLMs} without fine-tuning as baselines. For baselines with soft-thinking, we implement the soft-thinking reasoning pattern with the Gumbel-Softmax technique, using the best temperature setting $\tau_g=0.5$ proposed in \citet{wu2025llms}. Under both patterns, we follow the parameters in \citep{wu2025llms}, setting the temperature $\tau$ of LLMs to 0.6 (we discuss this setting in Appendix \ref{Appendixd2}), top-p to 0.95, and top-k to 30.

\textbf{Metrics.} We adopt the general metrics of Mean@32 (the average Pass@1 accuracy over 32 runs), Pass@16, and Pass@32, which measure the average probability of covering the correct answer within 16 and 32 runs, respectively.

\subsection{Main Result}

As shown in Table \ref{maindata}, SofT-GRPO can lead to a clear and consistent improvement from the No-Finetune results under the soft-thinking reasoning pattern. Moreover, SofT-GRPO outperforms applying soft-thinking on top of the discrete-token trained GRPO on average, demonstrating superior soft-thinking reasoning ability.

For Pass@1, SofT-GRPO consistently outperforms the discrete-token trained GRPO across LLMs of three sizes (+0.13\% on average accuracy). In contrast, comparing the +GRPO results on the discrete-token and soft-thinking patterns, performing soft-thinking on top of the discrete-token trained GRPO gives inconsistent improvement.

As observed in \citet{yue2025does}, GRPO can lead to decreased performance on Pass@K over the No-Finetune model, even though it improves performance for Pass@1. We can observe the same result in Table \ref{maindata}. In contrast, the SofT-GRPO leads to a clear improvement in the Pass@16 (+1.80\% on average) and Pass@32 (+2.19\% on average) metrics. Appendix \ref{Appendixd2} shows that the superiority will also hold for different temperatures. As detailed in Appendix \ref{Appendixe}, the improvement in Pass@16 and Pass@32 is likely due to the fact that for each sample in each LLM reasoning step, SofT-GRPO can reinforce the output probability of several tokens instead of focusing on one in GRPO.

\subsection{Comparison in Out-of-Domain Datasets}

Besides numerical reasoning, we conduct out-of-domain experiments to evaluate the general reasoning ability of SofT-GRPO. As shown in Table \ref{ood}, the LLMs fine-tuned with SofT-GRPO on numerical questions can still demonstrate advantages from Pass@1 to Pass@32 on a scientific reasoning benchmark (GPQA Diamond) and two code benchmarks (HumanEval and MBPP).

\vspace{-2pt}
\subsection{Boosting SofT-GRPO with Majority Voting}

To further exploit the advantage of SofT-GRPO on Pass@32, in this subsection, we design to boost SofT-GRPO with majority voting \citep{chen2024more}. As shown in Table \ref{majority voting}, SofT-GRPO with majority-voting can outperform No-Finetune LLM and LLM fine-tuned with discrete-token GRPO across Major@16 (the accuracy of the most common answer in 16 runs) and Major@32. The results show that SofT-GRPO-fine-tuned LLMs can be strengthened into better reasoning solvers with majority voting.

\begin{table}[t]
\centering
\renewcommand\arraystretch{1.08}
\setlength{\tabcolsep}{4mm}
\caption{Comparison of the proposed SofT-GRPO to the RLVR fine-tuning method proposed in \citep{butt2025soft}. 
We compare to their \textbf{reported results} under the same training dataset and LLM. Soft Tokens* represents their reported results fine-tuned under the soft-thinking pattern.}
\resizebox{0.49\textwidth}{!}{
\begin{tabular}{lcccc}
\bottomrule[0.5mm]
\multicolumn{1}{c|}{Dataset}    & \multicolumn{2}{c|}{MATH-500}  & \multicolumn{2}{c}{GSM8K} \\ \hline
\multicolumn{1}{c|}{Metrics}    & @1  & \multicolumn{1}{c|}{@32} & @1  & @32 \\ \hline\hline
\multicolumn{5}{c}{\textit{\textbf{LLaMA-3.2-3B-Instruct Base LLM}}} \\ \hline
\multicolumn{5}{c}{Discrete-Token CoT Reasoning Pattern}     \\ \hline
\multicolumn{1}{l|}{No-Finetune} & 38.1& \multicolumn{1}{c|}{84.0}& 67.9& 94.6\\
\multicolumn{1}{l|}{+ GRPO}     & 48.3& \multicolumn{1}{c|}{82.6}& 79.6& 96.5\\ \hline
\multicolumn{5}{c}{Soft-Thinking Reasoning Pattern}  \\ \hline
\multicolumn{1}{l|}{+ Soft Tokens*}    & 41.3& \multicolumn{1}{c|}{77.9}& 75.5& 95.2\\
\multicolumn{1}{l|}{}& 47.2  & \multicolumn{1}{c|}{83.4}  & 77.6  & 97.7  \\
\multicolumn{1}{l|}{\multirow{-2}{*}{+ SofT-GRPO}} & {\color[HTML]{588E31} (+5.9)} & \multicolumn{1}{c|}{{\color[HTML]{588E31} (+5.5)}} & {\color[HTML]{588E31} (+2.1)} & {\color[HTML]{588E31} (+2.5)} \\  \toprule[0.5mm]
\end{tabular}
}\vspace{-10pt}
\label{ablation-main}
\end{table}

\vspace{-2pt}
\subsection{Comparison to \citet{butt2025soft}}

We conduct comparison experiments with an existing RLVR algorithm for the soft-thinking pattern, the method in \citet{butt2025soft}. We refer to this method as Soft Tokens in Table \ref{ablation-main}, where the proposed SofT-GRPO can demonstrate a clear improvement compared to the results reported in \citet{butt2025soft}. Compared to SofT-GRPO, the algorithm in \citet{butt2025soft} also requires transitioning the $d$-dimensional vector between rollout workers and RLVR workers, making implementation more difficult. 

\begin{table*}[t]
\centering
\renewcommand\arraystretch{1.15}
\vspace{-2pt}
\setlength{\tabcolsep}{0.9mm}
\caption{Ablation studies on the noise added in the proposed SofT-GRPO. We highlight the performance differences on the ablation variants (i.e., adding Dirichlet noises or Gaussian noises).}
\resizebox{\textwidth}{!}{
\begin{tabular}{ccccccccccccccccccc}
\bottomrule[0.5mm]
\multicolumn{1}{c|}{Dataset}     & \multicolumn{3}{c|}{AIME2024}   & \multicolumn{3}{c|}{AIME2025} & \multicolumn{3}{c|}{AMC23}& \multicolumn{3}{c|}{MATH-500}  & \multicolumn{3}{c|}{GSM8K}     & \multicolumn{3}{c}{Average} \\ \hline
\multicolumn{1}{c|}{Metrics}     & @1     & @16    & \multicolumn{1}{c|}{@32}  & @1    & @16   & \multicolumn{1}{c|}{@32}  & @1     & @16& \multicolumn{1}{c|}{@32}& @1    & @16   & \multicolumn{1}{c|}{@32}   & @1    & @16   & \multicolumn{1}{c|}{@32}   & @1    & @16   & @32   \\ \hline\hline
\multicolumn{19}{c}{\textit{\textbf{DeepSeek-R1-Distill-Qwen-1.5B Base LLM}}}\\ \hline
\multicolumn{19}{c}{Soft-Thinking Reasoning Pattern} \\ \hline
\multicolumn{1}{l|}{No-Finetune}& 27.3   & 66.7   & \multicolumn{1}{c|}{70.0} & 23.8  & 46.7  & \multicolumn{1}{c|}{53.3} & 69.9   & 95.0   & \multicolumn{1}{c|}{95.0}   & 79.4  & 93.2  & \multicolumn{1}{c|}{96.6}  & 81.0  & 94.6  & \multicolumn{1}{c|}{97.1}  & 56.3  & 79.2  & 82.4  \\
\multicolumn{1}{l|}{+ GRPO}& 29.2   & 70.0   & \multicolumn{1}{c|}{73.3} & 25.4  & 46.7  & \multicolumn{1}{c|}{53.3} & 75.8   & 95.0   & \multicolumn{1}{c|}{95.0}   & 86.3  & 96.8  & \multicolumn{1}{c|}{98.2}  & 84.9  & 95.6  & \multicolumn{1}{c|}{96.4}  & 60.3  & 80.8  & 83.3  \\ \hline
\multicolumn{1}{l|}{+ SofT-GRPO (Original, Gumbel)}& 32.6   & 76.7   & \multicolumn{1}{c|}{80.0} & 26.1  & 50.0  & \multicolumn{1}{c|}{53.3} & 76.4   & 97.5   & \multicolumn{1}{c|}{97.5}   & 86.3  & 97.4  & \multicolumn{1}{c|}{98.0}  & 85.5  & 96.1  & \multicolumn{1}{c|}{97.0}  & 61.4  & 83.5  & 85.2  \\
\multicolumn{1}{c|}{}& 25.4   & 60.0   & \multicolumn{1}{c|}{70.0} & 22.9  & 56.7  & \multicolumn{1}{c|}{63.3} & 68.4   & 97.5   & \multicolumn{1}{c|}{97.5}   & 83.6  & 97.0  & \multicolumn{1}{c|}{97.8}  & 83.6  & 97.0  & \multicolumn{1}{c|}{97.8}  & 56.8  & 81.6  & 85.3  \\
\multicolumn{1}{l|}{\multirow{-2}{*}{+ SofT-GRPO (Dirichlet Noise)}} & {\color[HTML]{FF0000} (-7.2)}  & {\color[HTML]{FF0000} (-16.7)} & \multicolumn{1}{c|}{{\color[HTML]{FF0000} (-10)}} & {\color[HTML]{FF0000} (-3.2)} & {\color[HTML]{75BD42} (+6.7)} & \multicolumn{1}{c|}{{\color[HTML]{75BD42} (+10)}} & {\color[HTML]{FF0000} (-8)}    & {\color[HTML]{75BD42} (0)} & \multicolumn{1}{c|}{{\color[HTML]{75BD42} (0)}} & {\color[HTML]{FF0000} (-2.7)} & {\color[HTML]{FF0000} (-0.4)} & \multicolumn{1}{c|}{{\color[HTML]{FF0000} (-0.2)}} & {\color[HTML]{FF0000} (-1.8)} & {\color[HTML]{75BD42} (+0.9)} & \multicolumn{1}{c|}{{\color[HTML]{75BD42} (+0.8)}} & {\color[HTML]{FF0000} (-4.6)} & {\color[HTML]{FF0000} (-1.9)} & {\color[HTML]{75BD42} (+0.1)} \\
\multicolumn{1}{c|}{}& 21.9   & 60.0   & \multicolumn{1}{c|}{73.0} & 20.0  & 40.0  & \multicolumn{1}{c|}{43.3} & 65.6   & 97.5   & \multicolumn{1}{c|}{97.5}   & 79.9  & 95.2  & \multicolumn{1}{c|}{96.6}  & 81.8  & 94.5  & \multicolumn{1}{c|}{95.5}  & 53.8  & 77.4  & 81.2  \\
\multicolumn{1}{l|}{\multirow{-2}{*}{+ SofT-GRPO (Gaussian Noise)}}  & {\color[HTML]{FF0000} (-10.7)} & {\color[HTML]{FF0000} (-16.7)} & \multicolumn{1}{c|}{{\color[HTML]{FF0000} (-7)}}  & {\color[HTML]{FF0000} (-6.1)} & {\color[HTML]{FF0000} (-10)}  & \multicolumn{1}{c|}{{\color[HTML]{FF0000} (-10)}} & {\color[HTML]{FF0000} (-10.8)} & {\color[HTML]{75BD42} (0)} & \multicolumn{1}{l|}{{\color[HTML]{75BD42} (0)}} & {\color[HTML]{FF0000} (-6.5)} & {\color[HTML]{FF0000} (-2.2)} & \multicolumn{1}{c|}{{\color[HTML]{FF0000} (-1.4)}} & {\color[HTML]{FF0000} (-3.7)} & {\color[HTML]{FF0000} (-1.7)} & \multicolumn{1}{c|}{{\color[HTML]{FF0000} (-1.5)}} & {\color[HTML]{FF0000} (-7.6)} & {\color[HTML]{FF0000} (-6.1)} & {\color[HTML]{FF0000} (-4)}   \\ \toprule[0.5mm]
\end{tabular}
}
\vspace{-2pt}
\label{ablation2}
\end{table*}

\begin{figure*}[htbp]
    \centering
    \subfigure[Training reward curve of variants on added noises]{\includegraphics[width = 0.33\textwidth]{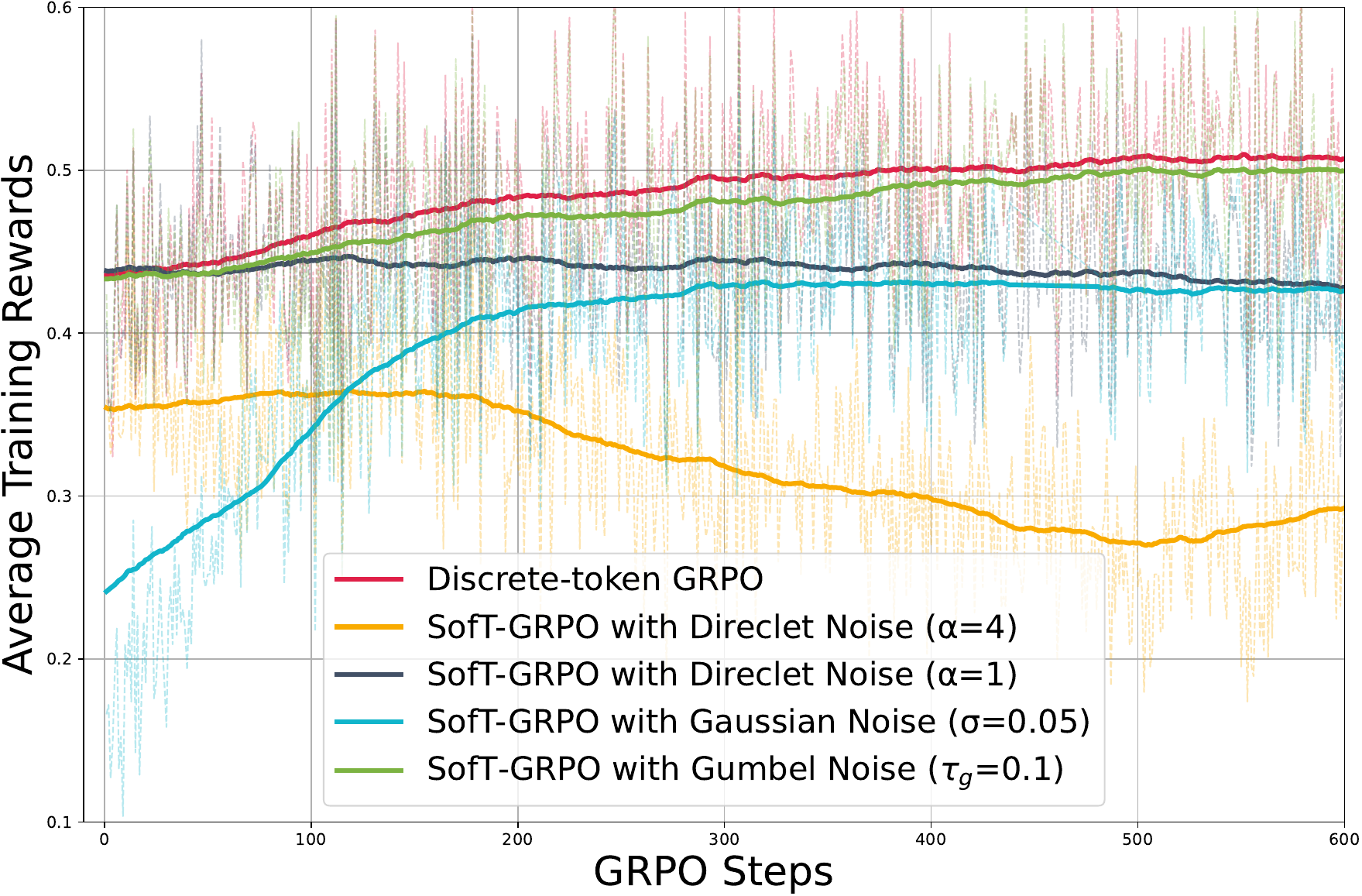}}
    \subfigure[Validation accuracy curve of variants on added noises]{\includegraphics[width = 0.33\textwidth]{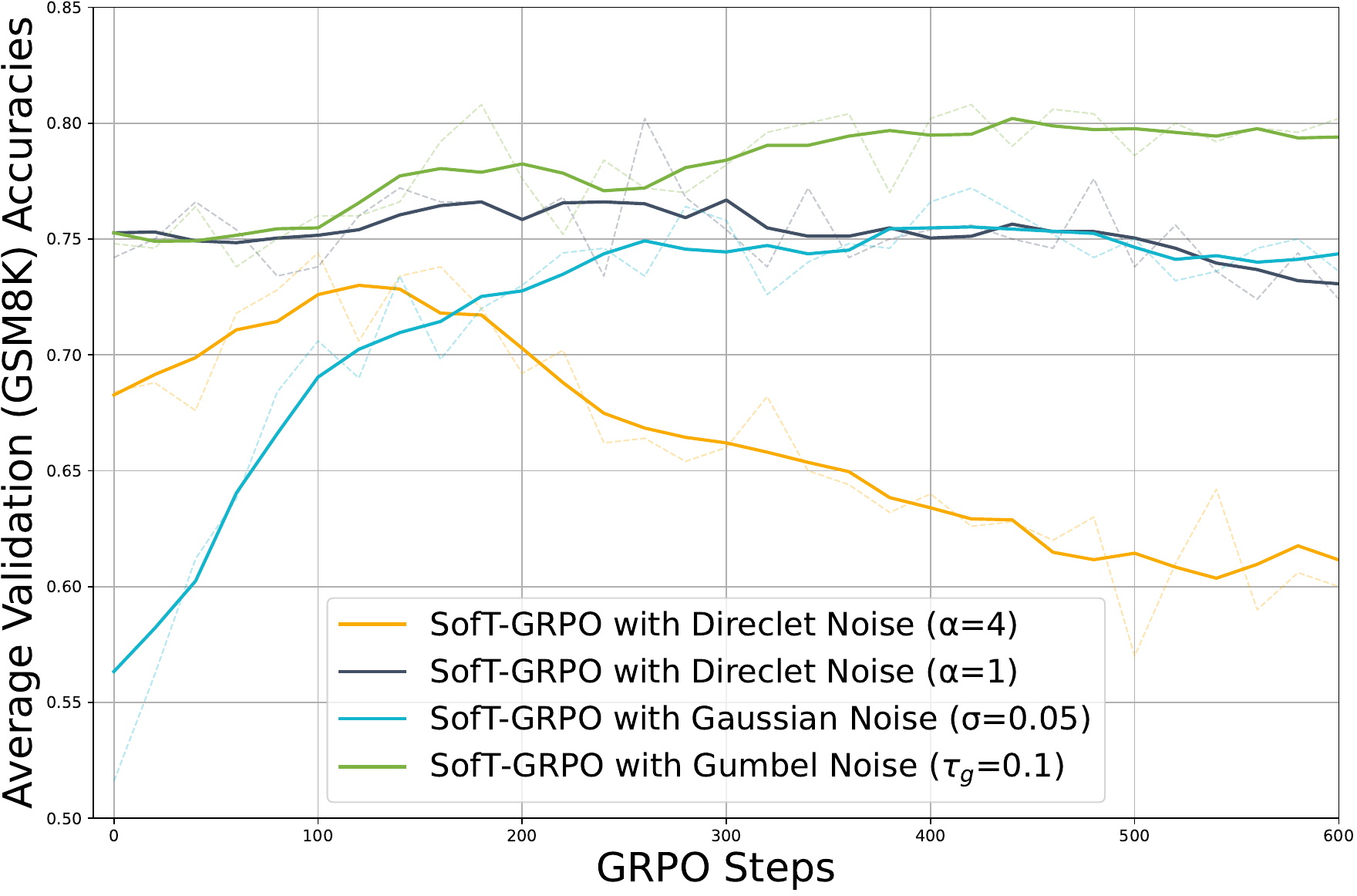}}
    \subfigure[Training reward curve of different hyper-parameters]{\includegraphics[width = 0.33\textwidth]{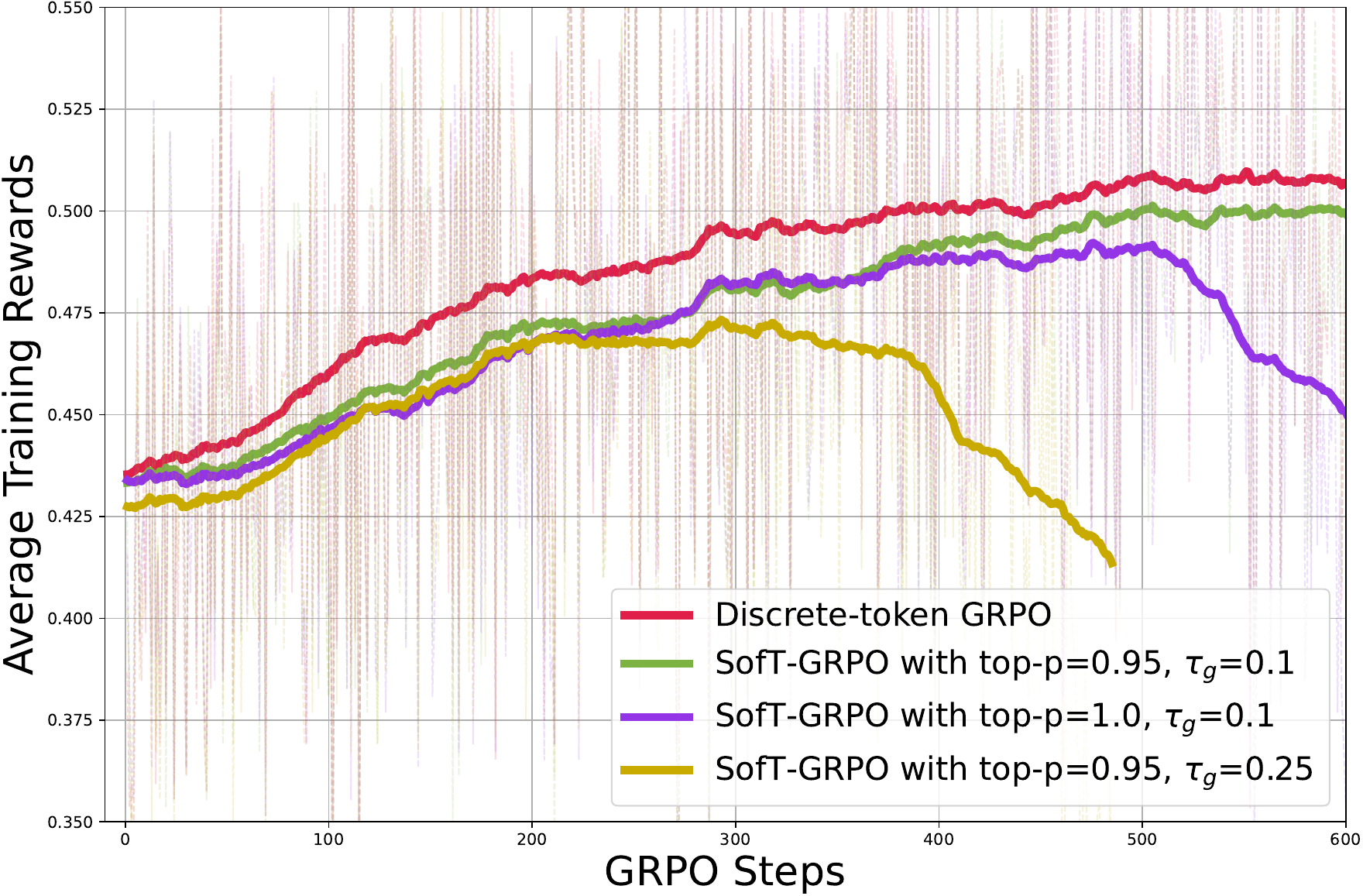}}\caption{Smoothed training or validation curves of ablation studies (the dashed background contains the data points). (a) discusses the setting of adding Gumbel noise in SofT-GRPO. (b,c) discusses the setting of top-p=0.95 and the Gumbel-Softmax temperature $\tau_g=0.1$.}\label{fig:curve1}
    \vspace{-2pt}
\end{figure*}

\vspace{-3pt}
\subsection{Comparison in the Token Efficiency}\label{main-tokencost}
In addition to accuracy, token efficiency is an important metric for LLMs. As shown in Appendix \ref{Appendixd1}, SofT-GRPO reduces thinking length compared to No-Finetune, and does not greatly increase tokens compared to discrete-token GRPO. Notably, SofT-GRPO leads to a clear reduction in thinking length for the \textit{LLaMA-3.2-3B-Instruct} model, highlighting its effectiveness in reducing computational costs.


\section{Discussion}


\subsection{Ablation on the Added Noise}

In SofT-GRPO, we employ the Gumbel-Softmax technique for controllable stochasticity. As discussed in Section \ref{priliminary-2}, adding Dirichlet noise can be another choice, so we compare the performance of the original SofT-GRPO and its two variants, adding Dirichlet noise or Gaussian noise to probabilities $p_i$ in training. The reward curve in training is shown in Figure \ref{fig:curve1}(a), the validation curve is shown in Figure \ref{fig:curve1}(b), and the final performance of variants is shown in Table \ref{ablation2}. Compared to adding the original Gumbel noises, LLMs do not gain improvements using Dirichlet noise, and adding the Gaussian noise will cause a poorer initial performance.

\subsection{Ablation on Hyper-Parameters}\label{hyperpara}
SofT-GRPO sets the top-p as 0.95 and the Gumbel temperature as $\tau_g=0.1$. To demonstrate the reason for these settings, in Figure \ref{fig:curve1}(c), we compare the training reward curve of the original SofT-GRPO with two variants (varying top-p to 1.0 or $\tau_g$ to 0.25). Results show that both variants will cause a collapse in the training process. As analyzed in Appendix \ref{Appendixd3}, the two variants will cause a substantial increase in the KL divergence between $\pi_\theta$ and $\pi_{\theta_{ref}}$. So, we attribute this kind of collapse to the fact that fine-tuning under the soft-thinking pattern may lead to soft-token inputs outside the pre-trained discrete-token embedding space.

\subsection{Visualization: Soft-Thinking after SofT-GRPO}

In Appendix \ref{Appendixg}, we provide a visualization of the soft-thinking reasoning path after SofT-GRPO fine-tuning. As shown in Figure \ref{fig:gsm8kexample}, high probability paths preserve interpretability, and we also observe new emerged distributions with high mass on a token as well as its antonym (Yeah and No), indicating preservation of multiple reasoning branches.

\section{Conclusion}
This paper presents a powerful RLVR algorithm, SofT-GRPO, to reinforce LLMs under the soft-thinking reasoning pattern. It integrates controllable stochasticity with Gumbel-Softmax and updates the soft-thinking policy with Gumbel reparameterization. It can demonstrate superior numerical, scientific, and code reasoning ability compared to the conventional discrete-token GRPO on Pass@1, especially Pass@16 and Pass@32. It can also be boosted into a better solver with majority voting. This work demonstrates the advantages of soft-thinking in LLM reasoning. Future works include extending the method for models such as Vision Language Models.








\section*{Impact Statement}

This paper presents work whose goal is to advance the field of machine learning. There are many potential societal consequences of our work, none of which we feel must be specifically highlighted here.

\bibliography{example_paper}
\bibliographystyle{icml2025}

\newpage
\appendix
\onecolumn
\newpage
\section*{Appendix Contents}
\begin{enumerate}
    \item \hyperref[Appendixa]{\textbf{Related Work}} \dotfill \pageref{Appendixa}
        \begin{enumerate}
            \item RLVR on Discrete-Token CoT Reasoning
            \item Latent Reasoning
            \item The Working Mechanism of Soft-thinking
        \end{enumerate}
    \item \hyperref[Appendixb]{\textbf{Prompt of Language Reasoning and Latent Generation}} \dotfill \pageref{Appendixb}
    \item \hyperref[Appendixc]{\textbf{Motivation and Theoretical Proof}} \dotfill \pageref{Appendixc}
        \begin{enumerate}
            \item \hyperref[Appendixc1]{Motivation: Mismatch in \citet{butt2025soft}} \dotfill \pageref{Appendixc1}
            \item \hyperref[Appendixc2]{Proof for Theorem 3.1} \dotfill \pageref{Appendixc2}
            \item \hyperref[Appendixc3]{Derivation of Eq.~\eqref{soft-grpo}, Eq.~\eqref{repa} and Eq.~\eqref{repac}} \dotfill \pageref{Appendixc3}
        \end{enumerate}
    \item \hyperref[Appendixc2]{\textbf{Detailed Parameters for Training \& Testing}} \dotfill \pageref{Appendixc2}
    \item \hyperref[Appendixd]{\textbf{Supplementary of Experiments}} \dotfill \pageref{Appendixd}
        \begin{enumerate}
            \item Supplementary of Token Efficiency
            \item Experiments under Different Temperatures
            \item Supplementary Ablation on Hyper-Parameters
            \item Pass@K up to K=1024
            \item P-values for Significance on Pass@K
        \end{enumerate}
    \item \hyperref[Appendixe]{\textbf{Discussion \& Analysis on Results}} \dotfill \pageref{Appendixe}
    \item \hyperref[Appendixf]{\textbf{Baselines \& Datasets \& Licenses}} \dotfill \pageref{Appendixf}
        \begin{enumerate}
            \item Baselines
            \item Datasets
            \item Inference Framework
            \item Licenses
        \end{enumerate}
    \item \hyperref[Appendixg]{\textbf{Visualization}} \dotfill \pageref{Appendixg}
\end{enumerate}
\newpage

\section{Related Work}\label{Appendixa}
In this section, we will discuss recent developments on discrete-token RLVR, a similar domain of soft-thinking, the latent reasoning methods, and the working mechanism of soft-thinking.

\subsection{RLVR on Discrete-Token CoT Reasoning}\label{Appendixa1}
Deepseek-R1-zero \citep{guo2025deepseek} has demonstrated remarkable performance with discrete-token RLVR. And there has been a wide collection of RLVR methods for discrete-token CoT policy optimization, such as GRPO \citep{shao2024deepseekmath}, Dr. GRPO \citep{liu2025understanding}, DAPO \citep{yu2025dapo}, and Lite PPO \citep{liu2025part}. To further improve these algorithms towards some specific goals, recent works focus on developing RLVR methods for efficient responses \citep{feng2025efficient} or controlling the entropy in CoT generation for better exploration-exploitation balance \citep{hao2025rethinking}. 

To generate concise discrete-token CoTs, \citet{arora2025training} modifies the reward function by adding penalties on the generation length. \citet{qi2025optimizing} and \citet{dai2025s} turn to formulate the fine-tuning process for concise language CoT as a multi-objective optimization task with the trade-off between token efficiency and accuracy. They truncate the generation of discrete-token CoT at several thinking budgets and optimize the overall performance over them.

RLVR methods with entropy control try to make a better balance between exploration and exploitation. They try to control the entropy in the training process to stimulate the exploration within the groups of CoTs in GRPO \citep{wang2025arbitrary}.

This paper only considers GRPO as the only RLVR pipeline, because we believe these modifications (e.g., considering entropy or DAPO, which proposes to remove the KL divergence term) should be orthogonal to the two major contributions of SofT-GRPO: Sampling with Gumbel-Softmax and credit assignment with Gumbel reparameterization. In the future, we would like to do a comprehensive investigation to check this.

\subsection{Latent Reasoning}\label{Appendixa2}
Similar to the soft-thinking pattern, latent reasoning methods pass continuous vectors between LLM steps. These methods fully decouple the reasoning process from explicit natural language (which soft-thinking does not do) and perform inference in the hidden space of the model. Generally, latent CoT methods are often diverse from each other and can be mainly divided into token-wise auto-regressive methods and auxiliary strategies \citep{chen2025latentcot,zhu2025survey}. Token-wise auto-regressive methods transform the reasoning process into 'soft thoughts' with dense latent embeddings \citep{hao2024training} or specialized tokens (e.g., pause \citep{goyal2024pause,zelikman2024quietstar}). These methods focus on transferring the original reasoning policy in the language domain to a latent embedding space, including curriculum learning (e.g., Coconut \citep{hao2024training}, LightThinker \citep{zhang2025lightthinker}, SIM-COT \citep{wei2025sim}), self-distillation (e.g., CODI \citep{shen2025codi}), and one-shot compression (e.g., CoLaR \citep{tan2025think}, SynAdapt \citep{wang2025synadapt}, Latent-SFT \cite{deng2025latent}). Auxiliary strategies (e.g., SoftCoT \citep{xu2025softcot}) generate latent embeddings from an auxiliary module and inject them into the frozen main model \citep{cheng2024compressed,xu2025softcot++,su2025token}.

Due to the existing token-wise auto-regressive methods completely treating the language CoTs as the label, these methods can hardly surpass or even reach the performance level of reasoning LLMs. So empirically, these methods can effectively improve the token efficiency compared to language CoT, but there is a clear performance drop. Auxiliary strategies, instead, can effectively boost the performance of the original LLM, sacrificing the running efficiency.

Compared to latent reasoning methods mainly aiming at better efficiency, the proposed SofT-GRPO aims at reinforcing the accuracy of the soft-thinking pattern to surpass discrete-token CoT with GRPO on general reasoning tasks.

\subsection{The Working Mechanism of Soft-thinking}\label{Appendixa3}

As illustrated in \citep{zhang2025soft}, soft-thinking is effective because it transforms the exponential path expansion of reasoning into a tractable and parallelizable process, rigorously approximating full path-summation without sacrificing efficiency or accuracy.

\begin{itemize}
    \item \textbf{Continuous Concept Space:} Soft Thinking replaces discrete token choices with ``concept tokens,'' which are probability-weighted mixtures over all token embeddings. This enables the model to represent more abstract and nuanced concepts beyond individual tokens.
    
    \item \textbf{Parallel Path Exploration:} By retaining the full probability distribution at each reasoning step, Soft Thinking allows the model to implicitly explore multiple reasoning trajectories in parallel, rather than committing early to a single discrete path.
\end{itemize}

Remarkably, LLMs can directly interpret and process soft tokens without the need for additional fine-tuning. Intuitively, this is because soft tokens are constrained to a well-confined space, specifically the \textbf{convex hull} formed by the model's embedding vectors.

\begin{itemize}
\item
\textbf{First-Order Linearization for Effectiveness:} 
Moreover, the effectiveness of the soft-thinking paradigm can be understood as a recursive first-order (linear) approximation of the full discrete autoregressive path-sum as follows:
\begin{enumerate}
    \item For discrete reasoning tokens $\boldsymbol{R} = (r_1, \ldots, r_{|R|})$, answer $\boldsymbol{A} = (a_1, \ldots, a_{|A|})$, and LLM policy $\pi_\theta$, we have:
    \begin{equation}
        \label{eq:full-path}
        p(\boldsymbol{A} \mid \boldsymbol{Q}) = 
        \sum_{R} 
        \prod_{t=1}^{|\boldsymbol{R}|}\pi_\theta(r_t \mid [\boldsymbol{Q}, (r_1, ..., r_{t-1})])
        \prod_{t=1}^{|\boldsymbol{A}|}\pi_\theta(a_t \mid [\boldsymbol{Q}, \boldsymbol{R}, (a_1, ..., a_{t-1})]).
    \end{equation}

    \item \textbf{Soft-Thinking Tokens:} At each reasoning step, define the soft token as the expected embedding:
    \begin{equation}
        \label{eq:soft-token}
        p_t \sim \pi_\theta(\cdot \mid [\boldsymbol{Q}, (\boldsymbol{s}_1, ..., \boldsymbol{s}_{t-1})]), \quad 
        \boldsymbol{s}_t = \sum_{i=1}^{|\mathcal{T}|} p_{t,i} \cdot e_i,
    \end{equation}
    where $p_t$ is the token probability distribution, $e_i$ is the embedding of token $i$.

    \item Then we can recursively replace each sum over discrete tokens by soft tokens. For example, for the first reasoning step:
    \begin{align}
        \label{eq:lin1}
        p(\boldsymbol{A}|\boldsymbol{Q}) 
        &= \sum_{r_1} \pi_\theta(r_1|\boldsymbol{Q}) \, 
            p(\boldsymbol{A}|\boldsymbol{Q}, r_1) \\
        &\approx 
            p(\boldsymbol{A}|\boldsymbol{Q},\, \mathbb{E}_{r_1}[e_{r_1}]) \\
        &= p(\boldsymbol{A} | \boldsymbol{Q},\, \boldsymbol{s}_1).
    \end{align}
    Here, $\boldsymbol{s}_1 = \sum_{i=1}^{|\mathcal{T}|} \pi_\theta(r_1=i|\boldsymbol{Q}) \cdot e_i$ is the soft token for the first step.

    For later steps, recursively apply the same procedure:
    \begin{align}
        p(\boldsymbol{A}|\boldsymbol{Q}, \boldsymbol{s}_1) &= 
        \sum_{r_2} \pi_\theta(r_2|\boldsymbol{Q}, \boldsymbol{s}_1) \, p(\boldsymbol{A}|\boldsymbol{Q}, \boldsymbol{s}_1, r_2) \\
        &\approx p(\boldsymbol{A}|\boldsymbol{Q}, \boldsymbol{s}_1, \boldsymbol{s}_2), \\
        \text{where} \ \boldsymbol{s}_2 &= \sum_{i=1}^{|\mathcal{T}|} \pi_\theta(r_2=i | \boldsymbol{Q}, \boldsymbol{s}_1) \cdot e_i.
    \end{align}

    \item \textbf{Recursive Soft-Token Approximation:}
    By recursively applying this linearization at each step, we approximate the original intractable path sum by a deterministic soft-token chain:
    \begin{equation}
        \label{eq:soft-path}
        p(\boldsymbol{A} \mid \boldsymbol{Q}) \approx 
        \prod_{t=1}^{|\boldsymbol{S}|} \pi_\theta (\boldsymbol{s}_t \mid [\boldsymbol{Q}, (\boldsymbol{s}_1, ..., \boldsymbol{s}_{t-1})])
        \prod_{t=1}^{|\boldsymbol{A}|} \pi_\theta(a_t \mid [\boldsymbol{Q}, \boldsymbol{S}, (a_1, ..., a_{t-1})]).
    \end{equation}
    where $\boldsymbol{S} = (\boldsymbol{s}_1, ..., \boldsymbol{s}_{|\boldsymbol{S}|})$. Thus, each $\boldsymbol{s}_t$ replaces the sum over all possible $r_t$, acting as a linear surrogate.
\end{enumerate}
\end{itemize}

However, it is important to note that not all points in the embedding space (i.e., \textbf{convex hull} of token embeddings) are equally interpretable or accessible to the LLM. For particularly smooth or spread-out distributions (making the approximation in Eq. \eqref{eq:lin1} fail), the expressiveness of soft tokens remains limited. We believe this limitation is closely related to the ``collapse'' phenomenon discussed in Section \ref{hyperpara}, where the model's reasoning may degrade or repeat when encountering out-of-distribution soft tokens.



\newpage
\section{Prompt of Language Reasoning and Latent Generation}\label{Appendixb}

In this part, we show the prompt we adopt for reasoning problems, including the in-domain numerical reasoning, out-of-domain GPQA reasoning, and code reasoning. We inherit the prompts in \citet{zhang2025soft} for out-of-domain benchmarks. We use the same prompt for both the soft-thinking reasoning pattern and the discrete-token reasoning pattern.

\begin{dialogbox}[Prompt for Numerical Reasoning.]
\textcolor{Green}{user}

\textcolor{RoyalBlue}{\{Question\}} Let's think step by step and output the final answer within \textbackslash boxed\{\}

\end{dialogbox}

\begin{dialogbox}[Prompt for GPQA Reasoning.]
\textcolor{Green}{user}

Please solve the following multiple-choice question. Please show your choice in the answer field with only the choice letter, e.g.,"answer": "C".

\textcolor{RoyalBlue}{\{Question\}}

\end{dialogbox}

\begin{dialogbox}[Prompt for Code Reasoning (HumanEval).]
\textcolor{Green}{user}

Please solve the programming task below in Python. Code should be wrapped in a markdown code block.

```python

\textcolor{RoyalBlue}{\{Question\}}

```

\end{dialogbox}

\begin{dialogbox}[Prompt for Code Reasoning (MBPP).]
\textcolor{Green}{user}

Please solve the programming task with test cases below in Python. Make sure your code satisfies the following requirements:

1. The function name and signature must match exactly as specified in the test cases.

2. Your code should be wrapped in a markdown code block without including any test cases.

Task:

\textcolor{RoyalBlue}{\{Question\}}

Test Cases:

```python

\textcolor{Brown}{\{TestCases\}}

```

\end{dialogbox}

The \textcolor{RoyalBlue}{blue} part represents the specific question (query $\boldsymbol{Q}$), the \textcolor{brown}{brown} part represents the possible test cases provided in the MBPP code reasoning benchmark. We use the code provided in \citet{zhang2025soft} in the verification process of the responses.

\newpage
\section{Motivation and Theoretical Proof}\label{Appendixc}

\subsection{Motivation: Mismatch in \citet{butt2025soft}}\label{Appendixc1}

\citet{butt2025soft} adds Gaussian noise on the soft-token inputs as follows:
\begin{equation}
    \boldsymbol{s}_t = \sum_{i=1}^{|\mathcal{T}|}p_i \cdot \boldsymbol{e}_i, \quad\hat{\boldsymbol{s}}_t  = \boldsymbol{s}_t +\mathcal{N}(0,\sigma^2 Id).
\end{equation}
and calculates the log probability as follows:
\begin{equation}
\log p(\hat{\boldsymbol{s}}_t) = -\frac{1}{2\sigma^2}||\hat{\boldsymbol{s}}_t-\boldsymbol{s}_t||^2_2 + \text{Constant}.
\end{equation}
Let $\boldsymbol{E} \in \mathbb{R}^{|\mathcal{T}| \times d}$ be the embedding matrix, and each token probability vector $\boldsymbol{p} \in \Delta^{|\mathcal{T}|-1}$ corresponds to the soft input $\boldsymbol{s} = \boldsymbol{E}^\top \boldsymbol{p} \in \mathbb{R}^{d}$. 

Assume the observed noisy soft input is $\hat{\boldsymbol{s}}$, and we define a likelihood
\begin{equation}
    \log p(\hat{\boldsymbol{s}}\mid \boldsymbol{p}) \propto -\frac{1}{2\sigma^2} \|\hat{\boldsymbol{s}}-\boldsymbol{E}^\top \boldsymbol{p}\|_2^2.
\end{equation}
Suppose we want to regard this as a likelihood on $\boldsymbol{p}$. In general, the mapping $\boldsymbol{p} \mapsto \boldsymbol{E}^\top \boldsymbol{p}$ is many-to-one: since $|\mathcal{T}| > d$, the kernel of $\boldsymbol{E}^\top$ is nontrivial, so
\begin{equation}
    \exists\, \boldsymbol{p}_1 \neq \boldsymbol{p}_2,\quad \text{with}\quad \boldsymbol{E}^\top \boldsymbol{p}_1 = \boldsymbol{E}^\top \boldsymbol{p}_2.
\end{equation}
Thus, the same $\boldsymbol{s}$ may correspond to infinitely many $\boldsymbol{p}$. This means that, under this Gaussian model, two different token mixtures can lead to the exact same log-probability value. The information about the original token distribution is partially lost in the embedding projection, unless $\boldsymbol{E}$ is invertible (which it is not). Thus, the use of a Gaussian noise model on the embedding space gives a mismatch to the true simplex-based probability geometry.

\textbf{In summary of Drawback 1:} Due to the non-injectivity (non-invertibility) of the embedding transformation, the model $\log p(\hat{\boldsymbol{s}}_t) \propto -\|\hat{\boldsymbol{s}}_t - \boldsymbol{s}_t\|^2$ does \emph{not} define a true likelihood on the simplex of token probabilities.

The above mismatch is not only due to the non-invertibility of the embedding matrix. Even if we restrict $\boldsymbol{p}$ to be sparse (nonzero only on a top-$k$ set, which is a general setting of LLMs or a common nature of LLM predictions), and even if the corresponding submatrix $\boldsymbol{E}_\mathcal{K}$ is invertible, the process of adding Gaussian noise in the $d$-dimensional embedding space fundamentally breaks the connection to sparse token distributions. 

More specifically, after adding Gaussian noise, with a general top-k setting (e.g., $k$=10 to 30, and $k<<d$) the perturbed embedding $\hat{\boldsymbol{s}}_t = \boldsymbol{s}_t + \boldsymbol{\epsilon}$ (with $\boldsymbol{\epsilon} \sim \mathcal{N}(0, \sigma^2 \boldsymbol{I}_d)$) will almost surely \textbf{not} lie in the convex hull of any set of $k$ token embeddings. In other words,
\begin{equation}
\forall~\hat{\boldsymbol{s}}_t,~\text{for almost all } \boldsymbol{\epsilon},~\nexists~k\text{-sparse } \boldsymbol{p} \text{ such that } \hat{\boldsymbol{s}}_t = \boldsymbol{E}^\top \boldsymbol{p}.
\end{equation}
The set of all top-$k$ soft-token embeddings forms a low-dimensional union of simplices in $\mathbb{R}^d$, which is a measure-zero subset of the space. The probability of a randomly perturbed embedding $\hat{\boldsymbol{s}}_t$ coinciding with a legal top-$k$ mixture is thus zero.

Therefore, defining the likelihood $p(\hat{\boldsymbol{s}}_t)$ as a simple Gaussian on embedding space cannot be interpreted as a likelihood on the space of top-$k$ soft tokens---not only due to non-invertibility or nonlinearity, but more fundamentally because most $\hat{\boldsymbol{s}}_t$ produced by noise are \emph{not} realizable by any top-$k$ soft-token distribution.

\textbf{In summary of Drawback 2:} Under the general top-k setting, the likelihood $p(\hat{\boldsymbol{s}}_t)$ as a simple Gaussian on the embedding space cannot be interpreted as a likelihood on the space of top-$k$ embeddings.

\newpage
\subsection{Proof for Theorem 3.1}\label{Appendixc2}

\noindent\textbf{Theorem \ref{theory31} (Gumbel-max Trick)}
\begin{itshape}
Let $(p_1, \dots, p_n)$ be nonnegative real numbers, not all zero. Let $g_1, \dots, g_n$ be independent samples from $\mathrm{Gumbel}(0, 1)$. Then,
\begin{equation}
    \Pr\left(j = \arg\max_{i}(g_i + \log  p_i)\right) = \frac{ p_j}{\sum_{i=1}^n  p_i}.
\end{equation}
\end{itshape}

\begin{proof}
For any $j \in \{1, \dots, n\}$,
\begin{align*}
    \Pr\left(j = \arg\max_{i}(g_i + \log  p_i)\right)
    &= \Pr\left( g_j + \log  p_j \geq g_i + \log  p_i, \,\, \forall i \neq j \right) \\
    &= \int_{-\infty}^{+\infty} \prod_{i \neq j} \Pr\left( g_i \leq g_j + \log  p_j - \log  p_i \right) f_{g_j}(g_j)\, dg_j,
\end{align*}
where $f_{g_j}(g) = e^{-g-\exp(-g)}$ is the PDF of the standard Gumbel distribution.

$\Pr(g_i \leq t) = F_\mathrm{Gumbel}(t) = \exp\left(-e^{-t}\right)$, so
\begin{align*}
    &\prod_{i \neq j} \Pr\left( g_i \leq g_j + \log  p_j - \log  p_i \right) \\
    &= \prod_{i \neq j} \exp \left( - e^{-(g_j + \log  p_j - \log  p_i)} \right) \\
    &= \exp \left( - \sum_{i \neq j} e^{-(g_j + \log  p_j - \log  p_i)} \right) \\
    &= \exp \left( - e^{-g_j} \sum_{i \neq j} \frac{ p_i}{ p_j} \right).
\end{align*}

The total probability is
\begin{align*}
    &\int_{-\infty}^{+\infty} e^{-g_j} e^{-e^{-g_j}} \exp \left( - e^{-g_j} \sum_{i \neq j} \frac{ p_i}{ p_j} \right) dg_j \\
    &= \int_{-\infty}^{+\infty} e^{-g_j} \exp \left( - e^{-g_j} \left[ 1 + \sum_{i \neq j} \frac{ p_i}{ p_j} \right] \right) dg_j.
\end{align*}

Let $S = 1 + \sum_{i \neq j} \frac{ p_i}{ p_j} = \frac{\sum_{i=1}^n  p_i}{ p_j}$. Substitute $y = e^{-g_j}, \, dg_j = -\frac{dy}{y}, \, y \in (0, +\infty)$,
\begin{align*}
    &= \int_{y=+\infty}^{0} y \exp(-y S)\left( -\frac{dy}{y} \right)
    = \int_{0}^{+\infty} \exp(-y S) dy
    = \frac{1}{S}
    = \frac{ p_j}{\sum_{i=1}^n  p_i}.
\end{align*}

Thus, combining all the above equations, 
\[
    \Pr\left(j = \arg\max_{i}(g_i + \log  p_i)\right) = \frac{ p_j}{\sum_{i=1}^n  p_i}.
\]
\end{proof}

\newpage
\subsection{Derivation of Eq.~\eqref{soft-grpo}, Eq.~\eqref{repa} and Eq.~\eqref{repac}}
\label{Appendixc3}

The Gumbel-Softmax rollout of SofT-GRPO is given in Eq. \eqref{soft-grporollout}. \textbf{On the determinism of the mapping.} It is crucial to note that, for fixed logits (or probabilities) $\mathbf{p}_t$ and Gumbel noises $\boldsymbol{\epsilon}_t$, the resulting soft token $\mathbf{s}_t$ is a \emph{deterministic function} of $(\mathbf{p}_t, \boldsymbol{\epsilon}_t)$:
\begin{equation}
\mathbf{s}_t = \mathrm{SoftEmbed}(\mathbf{p}_t, \boldsymbol{\epsilon}_t),\label{mapping}
\end{equation}
Hence, in our analysis, we always track the generative randomness via $\boldsymbol{\epsilon}_t$ (consider $\boldsymbol{\epsilon}_t$ as the action in RLVR), and \emph{not} via $\mathbf{s}_t$ directly. Cause every $\boldsymbol{g}'$ can map to one deterministic soft-token $\boldsymbol{s}_t$, but $\boldsymbol{s}_t$ can map to multiple $\boldsymbol{g}'$, calculating density on $\boldsymbol{g}'$ avoids the ill-posedness of attempting to define sampling densities directly in the soft-token space. So, we consider outputting $\boldsymbol{g}'$ as an action and put the generation of $\boldsymbol{s}_t$ as a part of the environment transition.

\paragraph{Eq.~\eqref{repa}: log-likelihood under the old policy.}
During rollout (generated by the old policy $\theta_{\mathrm{old}}$), we sample
Gumbel noises $\boldsymbol{\epsilon}_t=(\epsilon_{t,1},\dots,\epsilon_{t,|\mathcal{T}|})$.
The randomness of $\mathbf{s}_t$ is entirely attributable to $\boldsymbol{\epsilon}_t$, given $\mathbf{p}_t$. 
The joint density factorizes as
\[
p(\boldsymbol{\epsilon}_t)=\prod_{i=1}^{|\mathcal{T}|} f_{\mathrm{Gumbel}}(\epsilon_{t,i}),
\qquad
f_{\mathrm{Gumbel}}(x)=\exp\!\big(-x-\exp(-x)\big),
\]
where $f_{\mathrm{Gumbel}}$ represents the probability density function of a Gumbel distribution. 
Hence, the corresponding log-likelihood is
\[
\log p(\boldsymbol{\epsilon}_t)
=\sum_{i=1}^{|\mathcal{T}|}\log f_{\mathrm{Gumbel}}(\epsilon_{t,i})
=\sum_{i=1}^{|\mathcal{T}|}\Big[-\epsilon_{t,i}-\exp(-\epsilon_{t,i})\Big],
\]
which gives Eq.~\eqref{repa}. In this formulation, the density is over $\boldsymbol{\epsilon}_t$, while $\mathbf{s}_t$ is merely a deterministic function thereof.

\paragraph{Eq.~\eqref{repac}: log-likelihood under the current policy.}
For off-policy correction, we reuse the same realized perturbations from rollout.
Equivalently, we store the perturbed logits.
\[
g'_{t,i} \triangleq \log p^{\mathrm{old}}_{t,i}+\epsilon_{t,i},
\]
so that, under a new policy $\theta$ with probabilities $p_{t,i}=\pi_\theta(i\mid [Q,(\mathbf{s}_1,\dots,\mathbf{s}_{t-1})])$,
the implied noise that would have generated the same $g'_{t,i}$ under the \emph{new} policy is
\[
\epsilon_{t,i}=g'_{t,i}-\log p_{t,i}.
\]
Therefore, the log-density of the \emph{same} realized perturbation under the new policy is
\[
\sum_{i=1}^{|\mathcal{T}|}\log f_{\mathrm{Gumbel}}(g'_{t,i}-\log p_{t,i})
=
\sum_{i=1}^{|\mathcal{T}|}\Big[
-(g'_{t,i}-\log p_{t,i})
-\exp\!\big(-(g'_{t,i}-\log p_{t,i})\big)
\Big],
\]
which is exactly Eq.~\eqref{repac}. 
This preserves consistency in credit assignment and importance sampling,
as all stochasticity and probability mass are correctly attributed to 
$\boldsymbol{\epsilon}_t$ rather than $\mathbf{s}_t$.

\paragraph{Eq.~\eqref{soft-grpo}: Off-Policy Ratio} Cause the transformation from $\boldsymbol{g}'$ to $\boldsymbol{s}_t$ is fixed, the reparameterization process of SofT-GRPO considers the added noise as action. So, using the Eq.~\eqref{repa} and Eq.~\eqref{repac}, we can get the final result in Eq. \ref{soft-grpo} for the soft-thinking part.
\begin{equation}
    \begin{aligned}
 \frac{p(\boldsymbol{g}'| [\boldsymbol{Q}, (\boldsymbol{s}_{1}, \ldots, \boldsymbol{s}_{t-1})], \theta)}{p(\boldsymbol{g}'| [\boldsymbol{Q}, (\boldsymbol{s}_{1}, \ldots, \boldsymbol{s}_{t-1})], \theta_\text{old})}
        =&\frac{\exp\left(\sum_{i=1}^{|\mathcal{T}|}-(g'_i-\log p_i)-\exp(-g'_i+\log p_i)\right)}{\exp\left(\sum_{i=1}^{|\mathcal{T}|}-\epsilon_i-\exp(-\epsilon_i)\right)}\\
        =&\exp\left(\sum_{i=1}^{|\mathcal{T}|}\big(-(g'_i-\log p_i)-\exp(-g'_i+\log p_i)\big)-\big(-\epsilon_i-\exp(-\epsilon_i)\big)\right).
    \end{aligned}
\end{equation}

\newpage
\section{Detailed Parameters for Training \& Testing}\label{Appendixc2}

All our experiments are performed on 8 $\times$ H200 GPUs in about 2 to 3 days. Compared to discrete-token GRPO, SofT-GRPO requires $\times$2 to $\times$3 in time, mainly due to the more time consumption for soft-thinking rollout \citep{zhang2025soft, wu2025llm}. We show the experimental configurations in \ref{tab:nf} for No-Finetune, \ref{tab:grpo-para} for GRPO, and \ref{tab:soft-grpo-para} for SofT-GRPO.
\begin{table}[H]
    \centering
    \renewcommand\arraystretch{0.9}
    \caption{Parameters of No-Finetune}
    \small{\begin{tabularx}{0.8\textwidth}{
            l@{\hskip 0.8in}
            X
        }
    \toprule
         Parameter&  Value\\
         \midrule
                 \multicolumn{2}{c}{\textsc{No-Finetune Testing} under Discrete-Token Reasoning Paradigm} \\
        \midrule
        Maximum response length & $32768$ tokens \\
        Sampling temperature & 0.6 \\
        (top-p, top-k) & (0.95, 30) \\
         \midrule
                 \multicolumn{2}{c}{\textsc{No-Finetune Testing} under Soft-Thinking Reasoning Paradigm} \\
        \midrule
        Maximum response length & $32768$ tokens \\
        Sampling temperature & 0.6 \\
        Gumbel-temperature $\tau_g$ & 0.5 \\
        (top-p, top-k) & (0.95, 30) \\
         \bottomrule
    \end{tabularx}}
    \label{tab:nf}
\end{table}

\begin{table}[H]
    \centering
    \renewcommand\arraystretch{0.9}
    \caption{Parameters of Discrete-Token GRPO}
    \small{\begin{tabularx}{0.8\textwidth}{
            l@{\hskip 0.8in}
            X
        }
    \toprule
         Parameter&  Value\\
         \midrule
                 \multicolumn{2}{c}{\textsc{Discrete-Token GRPO Training}} \\
        \midrule
        Maximum response length & $8192$ tokens \\
        Sampling temperature & 1.0 \\
        (top-p, top-k) & (1, -1) \\
        Group Size $G$ & 8 \\
         Learning rate & $1\times 10^{-6}$ \\
         KL loss coefficient $\beta$ & 0.001 \\
         Policy clipping parameter $\epsilon$ & 0.2 \\
         \midrule
                 \multicolumn{2}{c}{\textsc{Discrete-Token GRPO Testing} under Discrete-Token Reasoning Paradigm} \\
        \midrule
        Maximum response length & $32768$ tokens \\
        Sampling temperature & 0.6 \\
        (top-p, top-k) & (0.95, 30) \\
         \midrule
                 \multicolumn{2}{c}{\textsc{GRPO Testing} under Soft-Thinking Reasoning Paradigm} \\
        \midrule
        Maximum response length & $32768$ tokens \\
        Sampling temperature & 0.6 \\
        Gumbel-temperature $\tau_g$ & 0.5 \\
        (top-p, top-k) & (0.95, 30) \\
         \bottomrule
    \end{tabularx}}
    \label{tab:grpo-para}
\end{table}

\begin{table}[H]
    \centering
    \renewcommand\arraystretch{0.9}
    \caption{Parameters of SofT-GRPO}
    \small{\begin{tabularx}{0.8\textwidth}{
            l@{\hskip 0.8in}
            X
        }
    \toprule
         Parameter&  Value\\
         \midrule
                 \multicolumn{2}{c}{\textsc{SofT-GRPO Training}} \\
        \midrule
        Maximum response length & $8192$ tokens \\
        Sampling temperature & 1.0 \\
        (top-p, top-k) & (0.95, 5) \\
        Group Size $G$ & 8 \\
        Gumbel-temperature $\tau_g$ & 0.1 \\
         Learning rate & $1\times 10^{-6}$ \\
         KL loss coefficient $\beta$ & 0.001 \\
         Policy clipping parameter $\epsilon$ & 0.2 \\
         \midrule
                 \multicolumn{2}{c}{\textsc{SofT-GRPO Testing} under Soft-Thinking Reasoning Paradigm} \\
        \midrule
        Maximum response length & $32768$ tokens \\
        Sampling temperature & 1.0 \\
        Gumbel-temperature $\tau_g$ & 0.1 \\
        (top-p, top-k) & (0.95, 5) \\
         \bottomrule
    \end{tabularx}}
    \label{tab:soft-grpo-para}
\end{table}

\newpage
\section{Supplementary of Experiments}\label{Appendixd}

\begin{table}[H]
\centering
\renewcommand\arraystretch{1.0}
\setlength{\tabcolsep}{2mm}
\caption{Experiments on the token efficiency for baselines and the proposed SofT-GRPO. \#Token values in Table represent the number of tokens across all queries, and the \#Token\_c values represent the number of tokens across correct queries.}
\resizebox{\textwidth}{!}{
\begin{tabular}{lcccccccccccc}
\bottomrule[0.5mm]
\multicolumn{1}{c|}{Dataset}& \multicolumn{2}{c|}{AIME2024} & \multicolumn{2}{c|}{AIME2025} & \multicolumn{2}{c|}{AMC23}  & \multicolumn{2}{c|}{MATH-500} & \multicolumn{2}{c|}{GSM8K}  & \multicolumn{2}{c}{Average} \\ \hline
\multicolumn{1}{c|}{Metrics}& \#Token & \multicolumn{1}{c|}{\#Token\_c} & \#Token & \multicolumn{1}{c|}{\#Token\_c} & \#Token & \multicolumn{1}{c|}{\#Token\_c} & \#Token & \multicolumn{1}{c|}{\#Token\_c} & \#Token & \multicolumn{1}{c|}{\#Token\_c} & \#Token    & \#Token\_c   \\ \hline\hline
\multicolumn{13}{c}{\textit{\textbf{DeepSeek-R1-Distill-Qwen-1.5B Base LLM}}} \\ \hline
\multicolumn{13}{c}{Discrete-Token CoT Reasoning Pattern}     \\ \hline
\multicolumn{1}{l|}{No-Finetune} & 16241.6  & \multicolumn{1}{c|}{14997.8}     & 16416.3  & \multicolumn{1}{c|}{13448.8}     & 10052.2  & \multicolumn{1}{c|}{9394.2}& 5616.6   & \multicolumn{1}{c|}{5368.1}& 1839.3   & \multicolumn{1}{c|}{1772.5}& 10033.2     & 8996.3\\
\multicolumn{1}{l|}{+ GRPO} & 8927.6   & \multicolumn{1}{c|}{8535.6}& 8039.4   & \multicolumn{1}{c|}{6414.2}& 5001.3   & \multicolumn{1}{c|}{4672.8}& 3106.1   & \multicolumn{1}{c|}{2958.5}& 1417.1   & \multicolumn{1}{c|}{1370.6}& 5298.3& 4790.3\\ \hline
\multicolumn{13}{c}{Soft-Thinking Reasoning Pattern}  \\ \hline
\multicolumn{1}{l|}{No-Finetune} & 17857.8  & \multicolumn{1}{c|}{16191.3}     & 17569.4  & \multicolumn{1}{c|}{14582.4}     & 11269.9  & \multicolumn{1}{c|}{10482.0}     & 4015.9   & \multicolumn{1}{c|}{3888.3}& 1699.3   & \multicolumn{1}{c|}{1649.9}& 10482.5     & 9358.8\\
\multicolumn{1}{l|}{+ GRPO} & 9383.2   & \multicolumn{1}{c|}{8934.5}& 8007.2   & \multicolumn{1}{c|}{7325.8}& 5203.7   & \multicolumn{1}{c|}{4894.4}& 3233.6   & \multicolumn{1}{c|}{3131.7}& 1385.5   & \multicolumn{1}{c|}{1349.9}& 5442.7& 5127.2\\
\multicolumn{1}{l|}{+ SofT-GRPO}  & 11039.6  & \multicolumn{1}{c|}{10756.1}     & 10519.6  & \multicolumn{1}{c|}{7831.3}& 5900.2   & \multicolumn{1}{c|}{5630.4}& 3549.5   & \multicolumn{1}{c|}{3399.2}& 1577.6   & \multicolumn{1}{c|}{1542.5}& 6517.3& 5831.9\\ \hline\hline
\multicolumn{13}{c}{\textit{\textbf{LLaMA-3.2-3B-Instruct Base LLM}}} \\ \hline
\multicolumn{13}{c}{Discrete-Token CoT Reasoning Pattern}     \\ \hline
\multicolumn{1}{l|}{No-Finetune} & 5010.5   & \multicolumn{1}{c|}{3334.1}& 4297.6   & \multicolumn{1}{c|}{2430.9}& 2790.7   & \multicolumn{1}{c|}{2398.1}& 1850.1   & \multicolumn{1}{c|}{1494.2}& 236.6    & \multicolumn{1}{c|}{224.6} & 2837.1& 1976.4\\
\multicolumn{1}{l|}{+ GRPO} & 8991.8   & \multicolumn{1}{c|}{6368.9}& 8443.8   & \multicolumn{1}{c|}{11088.3}     & 4998.1   & \multicolumn{1}{c|}{3978.5}& 3334.8   & \multicolumn{1}{c|}{2463.0}& 502.0    & \multicolumn{1}{c|}{453.6} & 5254.1& 4870.5\\ \hline
\multicolumn{13}{c}{Soft-Thinking Reasoning Pattern}  \\ \hline
\multicolumn{1}{l|}{No-Finetune} & 4859.2   & \multicolumn{1}{c|}{2727.7}& 4934.8   & \multicolumn{1}{c|}{3735.2}& 3259.9   & \multicolumn{1}{c|}{2624.0}& 2086.1   & \multicolumn{1}{c|}{1494.1}& 243.5    & \multicolumn{1}{c|}{227.8} & 3076.7& 2161.8\\
\multicolumn{1}{l|}{+ GRPO} & 9749.0   & \multicolumn{1}{c|}{7627.8}& 8498.9   & \multicolumn{1}{c|}{10041.4}     & 5389.9   & \multicolumn{1}{c|}{4987.3}& 3701.9   & \multicolumn{1}{c|}{2764.6}& 541.9    & \multicolumn{1}{c|}{505.8} & 5576.3& 5185.4\\
\multicolumn{1}{l|}{+ SofT-GRPO}  & 829.7    & \multicolumn{1}{c|}{862.9} & 893.1    & \multicolumn{1}{c|}{879.6} & 911.5    & \multicolumn{1}{c|}{927.9} & 632.8    & \multicolumn{1}{c|}{575.7} & 294.6    & \multicolumn{1}{c|}{292.9} & 712.3 & 707.8 \\ \hline\hline
\multicolumn{13}{c}{\textit{\textbf{DeepSeek-R1-Distill-Qwen-7B Base LLM}}}   \\ \hline
\multicolumn{13}{c}{Discrete-Token CoT Reasoning Pattern}     \\ \hline
\multicolumn{1}{l|}{No-Finetune} & 13120.7  & \multicolumn{1}{c|}{11511.4}     & 14347.7  & \multicolumn{1}{c|}{11750.7}     & 6346.0   & \multicolumn{1}{c|}{5987.1}& 3998.9   & \multicolumn{1}{c|}{3939.0}& 1061.1   & \multicolumn{1}{c|}{1032.0}& 7774.9& 6844.0\\
\multicolumn{1}{l|}{+ GRPO} & 7795.9   & \multicolumn{1}{c|}{7116.1}& 8003.5   & \multicolumn{1}{c|}{7369.8}& 4050.9   & \multicolumn{1}{c|}{3719.7}& 2473.3   & \multicolumn{1}{c|}{2405.5}& 1146.7   & \multicolumn{1}{c|}{1118.7}& 4694.0& 4346.0\\ \hline
\multicolumn{13}{c}{Soft-Thinking Reasoning Pattern}  \\ \hline
\multicolumn{1}{l|}{No-Finetune} & 13017.4  & \multicolumn{1}{c|}{11888.6}     & 14116.8  & \multicolumn{1}{c|}{11507.4}     & 6346.7   & \multicolumn{1}{c|}{6043.6}& 3947.4   & \multicolumn{1}{c|}{3858.6}& 996.3    & \multicolumn{1}{c|}{962.2} & 7684.9& 6852.1\\
\multicolumn{1}{l|}{+ GRPO} & 7464.9   & \multicolumn{1}{c|}{6531.4}& 8291.6   & \multicolumn{1}{c|}{7160.4}& 3931.7   & \multicolumn{1}{c|}{3736.7}& 2473.2   & \multicolumn{1}{c|}{2423.0}& 1117.0   & \multicolumn{1}{c|}{1104.5}& 4655.7& 4191.2\\
\multicolumn{1}{l|}{+ SofT-GRPO}  & 8035.8   & \multicolumn{1}{c|}{7556.8}& 8381.7   & \multicolumn{1}{c|}{7873.7}& 4008.0   & \multicolumn{1}{c|}{3843.5}& 2630.2   & \multicolumn{1}{c|}{2583.5}& 1293.2   & \multicolumn{1}{c|}{1276.4}& 4869.8& 4626.8\\ \toprule[0.5mm]
\end{tabular}
}
\label{length}
\end{table}

\subsection{Supplementary of Token Efficiency}\label{Appendixd1}

\begin{figure}[H]
    \centering
    \includegraphics[width=0.55\linewidth]{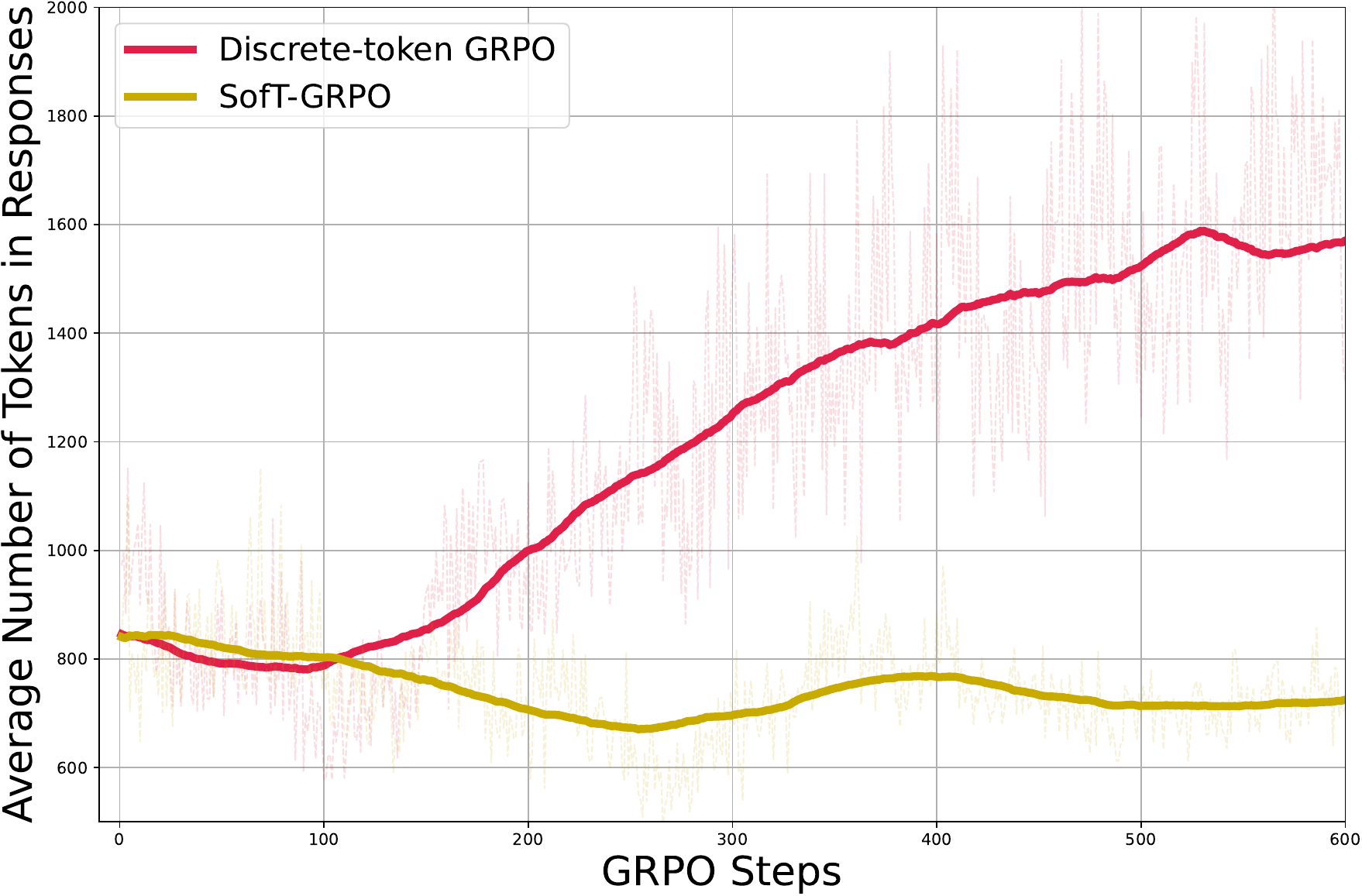}
    \caption{Token consumption curve on \textit{\textbf{LLaMA-3.2-3B-Instruct Base LLM}} during training.}
    \label{fig:llama-token}
\end{figure}

As shown in Section \ref{main-tokencost}, besides the performance, the token efficiency of LLMs is also an important metric. In this section, we compare the token efficiency of baselines and SofT-GRPO in Table \ref{length}. Compared to No-Finetune variants, SofT-GRPO can demonstrate a clear refinement in both the token efficiency across all queries and the token efficiency across correct queries. Compared to the discrete-token GRPO, SofT-GRPO will not cause severe token improvement. Specifically, we observe a severe reduction in the thinking length of the \textit{LLaMA-3.2-3B-Instruct} model. As shown in Figure \ref{fig:llama-token}, unlike GRPO, SofT-GRPO maintains and even enhances the token efficiency compared to the base LLM when training progresses, demonstrating the effectiveness of SofT-GRPO in reducing computational consumption.

\subsection{Experiments under Different Temperatures}\label{Appendixd2}
In Section \ref{experiment-section}, we adopt the setting of $\tau=0.6$ for discrete-token CoT. To investigate whether other temperature settings will break our observations in our main experiments, similar to \citet{yue2025does}, we try temperatures from the collection of $\tau=0.6$, $\tau=0.8$, $\tau=1.0$, $\tau=1.2$, and $\tau=1.4$. 

\begin{figure}[H]
    \centering
    \includegraphics[width=0.9\linewidth]{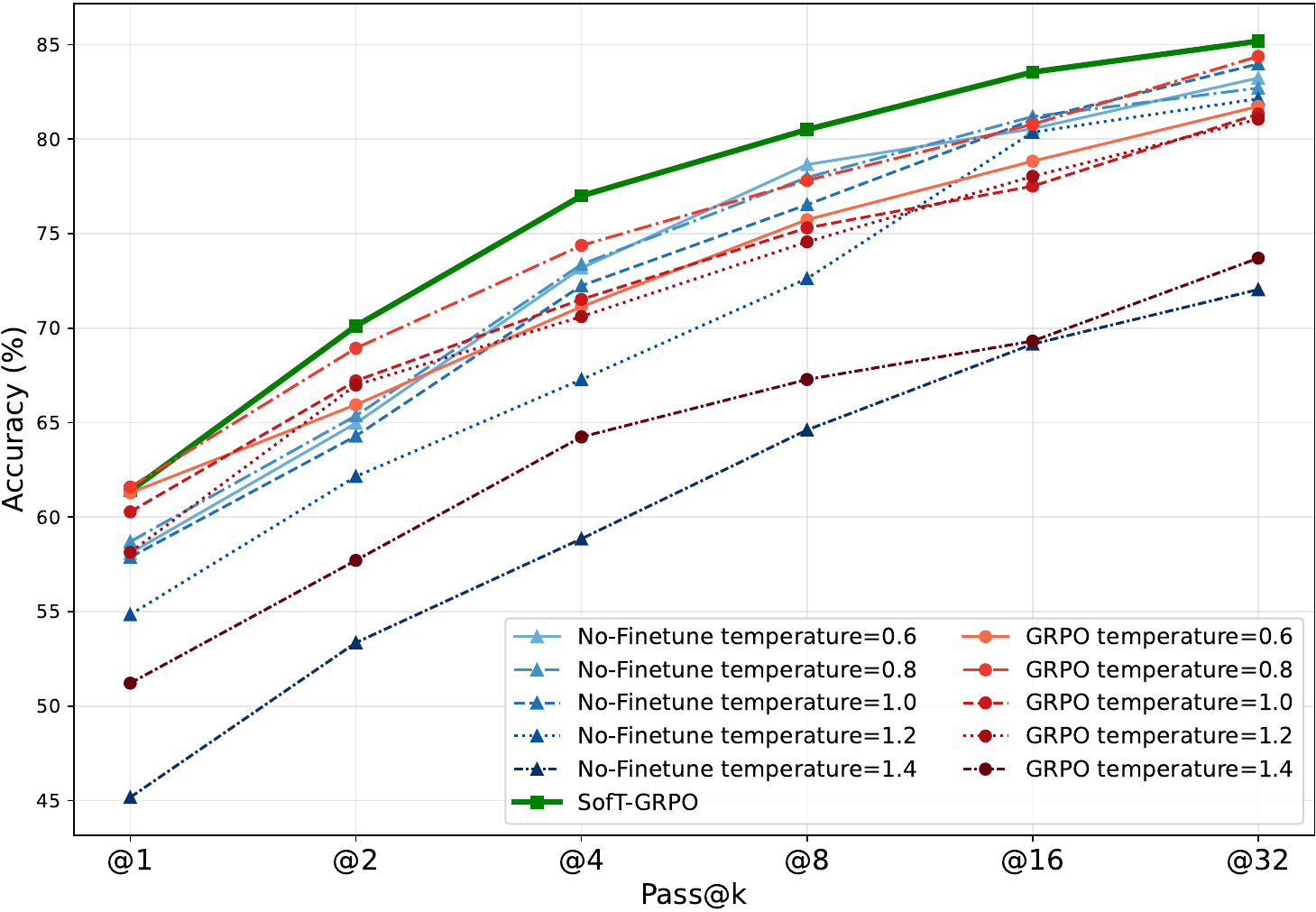}
    \caption{Running discrete-token CoT methods (GRPO and No-finetune) with more temperature options on \textit{\textbf{DeepSeek-R1-Distill-Qwen-1.5B Base LLM}}. Pass@K represents the pass rate within at most k runs, and Pass@1 is additionally averaged from 32 runs. Experiments are run on the five datasets in Table \ref{maindata} for the average. }
    \label{fig:temperature}
\end{figure}

As shown in Figure \ref{fig:temperature}, we conduct experiments on the 1.5B LLMs (i.e., \textit{\textbf{DeepSeek-R1-Distill-Qwen-1.5B Base LLM}}) for their average Pass@K accuracies across five numerical reasoning benchmarks (i.e., AIME2024, AIME2025, AMC23, MATH-500, GSM8K), where the proposed SofT-GRPO can demonstrate outstanding results compared to GRPO and No-Finetune variants with various temperatures, from Pass@1 to Pass@32.

\newpage
\subsection{Supplementary Ablation on Hyper-Parameters}\label{Appendixd3}

\begin{figure*}[htbp]
    \centering
    \subfigure[Training KL Divergence curve between $\pi_{\theta_\text{ref}}$ and $\pi_{\theta}$]{\includegraphics[width = 0.49\textwidth]{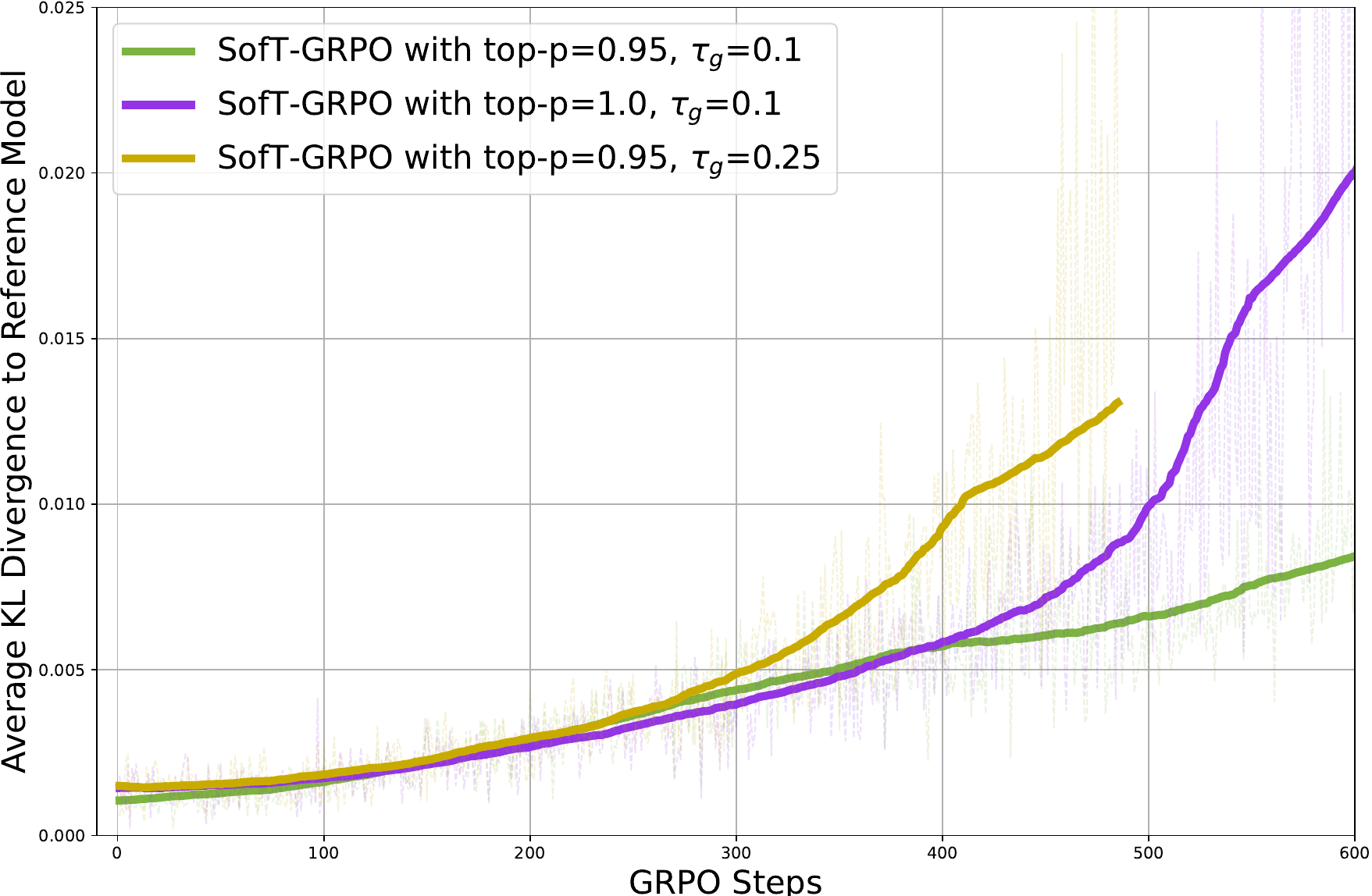}}
    \subfigure[Training Proximal Policy Optimization (PPO) \citep{schulman2017proximal} KL Divergence curve between $\pi_{\theta_\text{old}}$ and $\pi_{\theta}$]{\includegraphics[width = 0.49\textwidth]{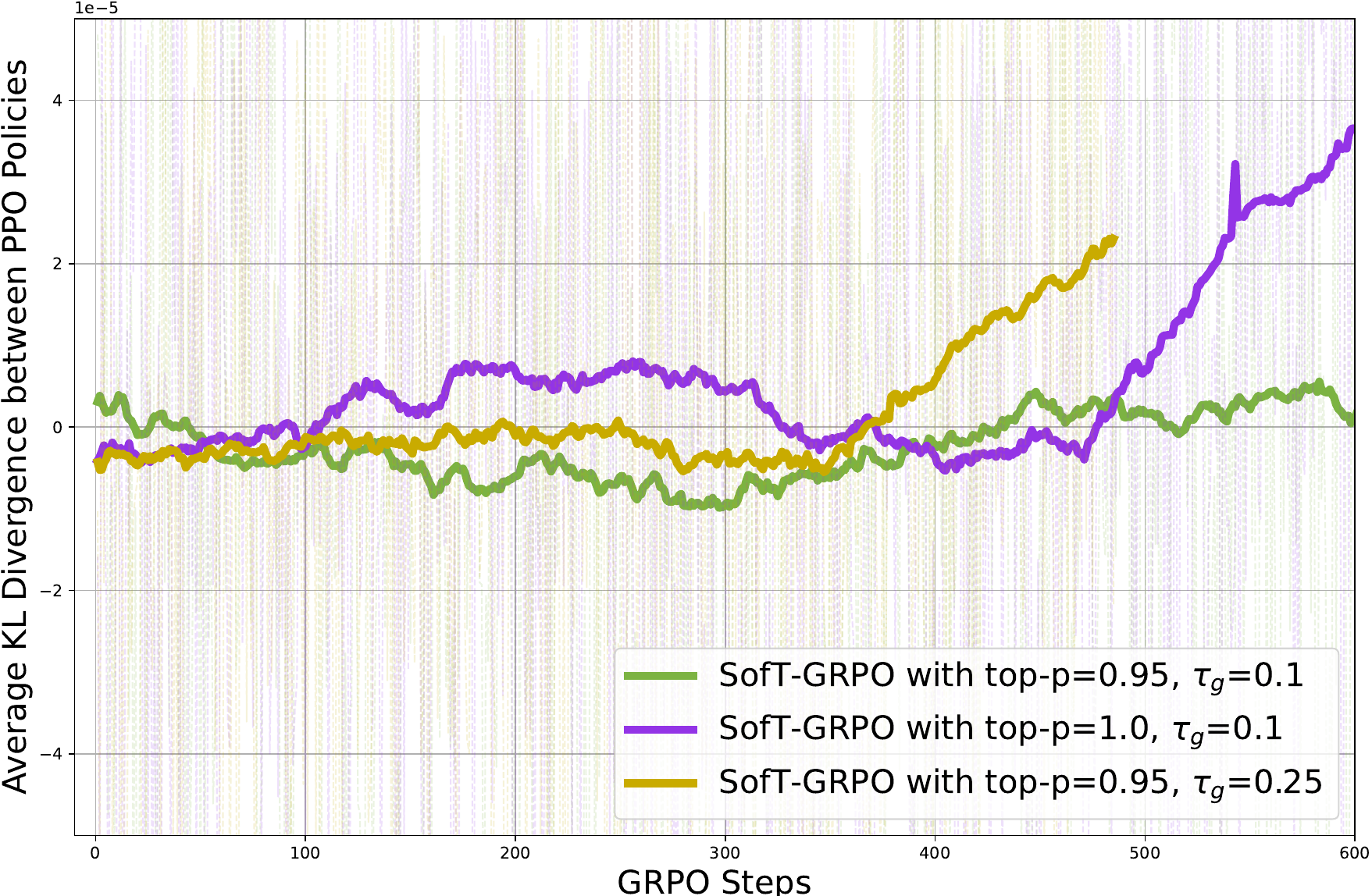}}\caption{KL Divergence curves in SofT-GRPO over \textit{\textbf{DeepSeek-R1-Distill-Qwen-1.5B}} Base LLM.}\label{fig:curve_kl}
\end{figure*}

In this subsection, we further investigate the hyperparameter settings. As briefly shown in Section \ref{hyperpara}, adopting a higher top-p or $\tau_g$ will cause collapses in training. We attribute these collapses to the case that some soft-thinking tokens may become incomprehensible for LLMs. 

As shown in Figure \ref{fig:curve_kl}(a), the variants (varying top-p to 1.0 or varying $\tau_g$ to 0.25) will demonstrate higher divergence between the fixed pre-trained $\pi_{\theta_\text{ref}}$ and the current policy, which can be an indicator of the inputs outside the pre-trained embedding space.

Recently, \citet{liu-li-2025, qi2025defeating} provides excellent insight into the collapse situation in the RLVR fine-tuning. When collapse is caused by a precision issue, they observe a very high KL divergence between $\pi_{\theta_\text{old}}$ and $\pi_{\theta}$ (Refer to Figure 3 in \citep{qi2025defeating}, $10^{-3}$ even higher). However, in the variants shown in Figure \ref{fig:curve_kl}(b), we find that their KL divergence in PPO policies is less than $10^{-5}$, indicating that the variants of SofT-GRPO are less likely to have precision issues.

\newpage

\subsection{Pass@K up to K=1024}

Table \ref{maindata} and Figure \ref{fig:temperature} show the Pass@K for K at most 64. In \citet{yue2025does}, they conduct experiments for K up to 1024. We adopt this setting. Confined by limited computational resources, we implement the proposed SofT-GRPO over \textbf{\textit{DeepSeek-R1-Distill-Qwen-1.5B}} for 1024 runs, as well as the GRPO and No-Finetune baselines with the discrete-token CoT reasoning pattern.

\begin{figure*}[htbp]
    \centering
    \subfigure[Pass@16 to Pass@1024 on AIME2024]{\includegraphics[width = 0.49\textwidth]{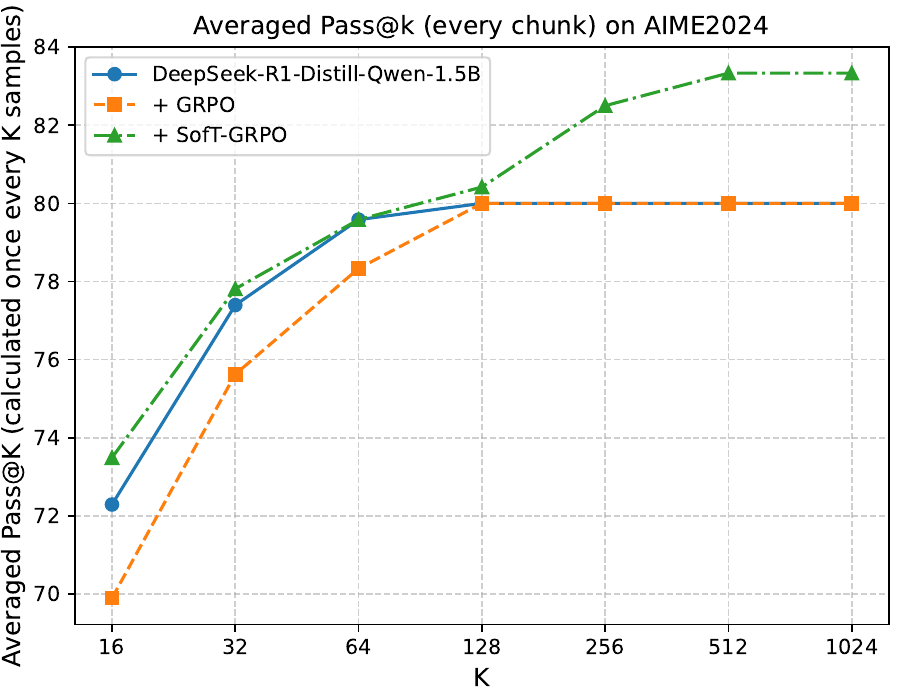}}
    \subfigure[Pass@16 to Pass@1024 on AMC23]{\includegraphics[width = 0.49\textwidth]{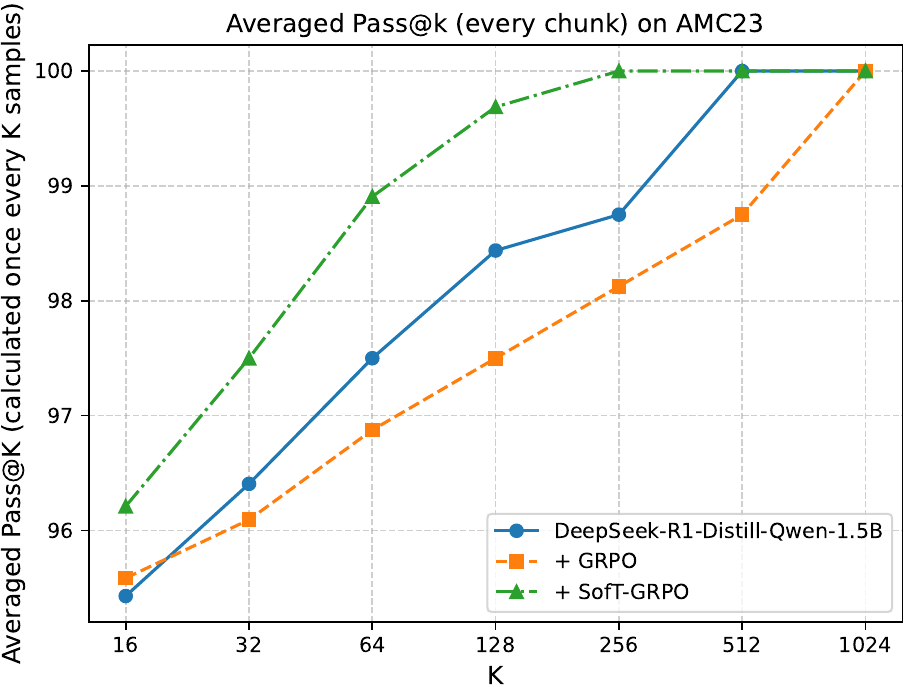}}\caption{Pass@16 to Pass@1024 on AIME2024 and AMC 23. The fluctuation in Figure (a) occurs because AIME only has 30 instances, and SofT-GRPO correctly solved three out of four (1024/256) problems when K=256. In Figure (b), all algorithms ultimately achieved complete correctness. Consistent with this, we checked CoT to prove that the correct results did not come from random guessing.}\label{fig:pass1024}
\end{figure*}

As shown in Figure \ref{fig:pass1024}, we calculate Pass@16 to Pass@1024 on AIME2024 and AMC 23. As a difference from the result in Table \ref{maindata}, \textbf{we obtain Pass@K for each K runs instead of the first K runs}. For example, when calculating Pass@32, we divide the 1024 runs into 32 groups, calculating their Pass@K value and averaging. In Table \ref{maindata} instead, we calculate the Pass@K at first 32 runs.

Figure \ref{fig:pass1024} obtains the same result pattern compared to the results in \citep{yue2025does}, where GRPO will outperform the No-finetune version when K is small, but will be left behind when K increases. SofT-GRPO can consistently outperform the GRPO and no-finetune counterparts on all K (from 16 to 1024).

\newpage

\subsection{P-values for Significance on Pass@K}
\begin{figure*}[htbp]
    \centering
    \subfigure[P-scores of Pass@16 over SofT-GRPO, GRPO, and No-finetune on AIME2024]{\includegraphics[width = 0.35\textwidth]{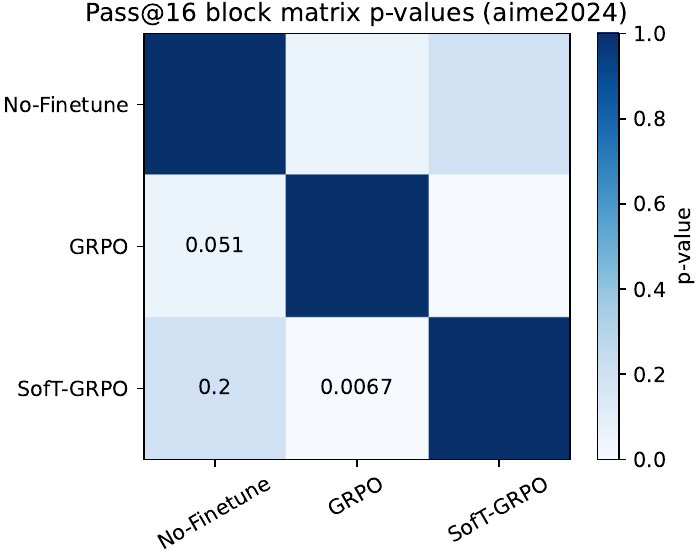}}\qquad\qquad
    \subfigure[P-scores of Pass@32 over SofT-GRPO, GRPO, and No-finetune on AIME2024]{\includegraphics[width = 0.35\textwidth]{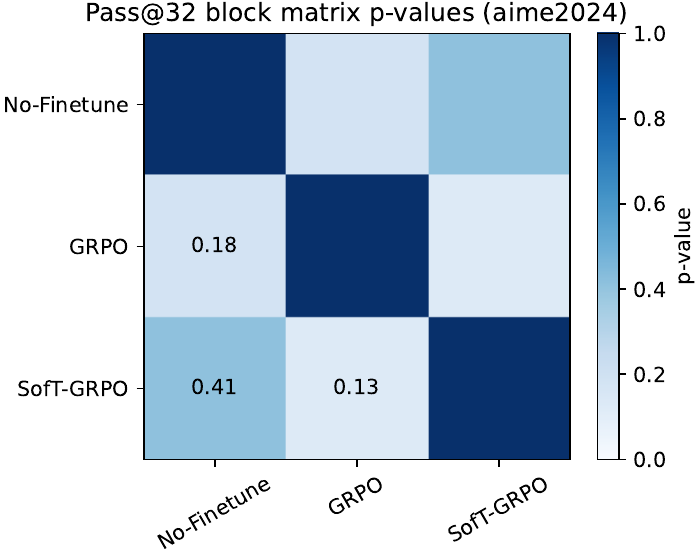}}\caption{Significance test on AIME2024.}\label{fig:paime}
\end{figure*}

\begin{figure*}[htbp]
    \centering
    \subfigure[P-scores of Pass@16 over SofT-GRPO, GRPO, and No-finetune on AMC23]{\includegraphics[width = 0.35\textwidth]{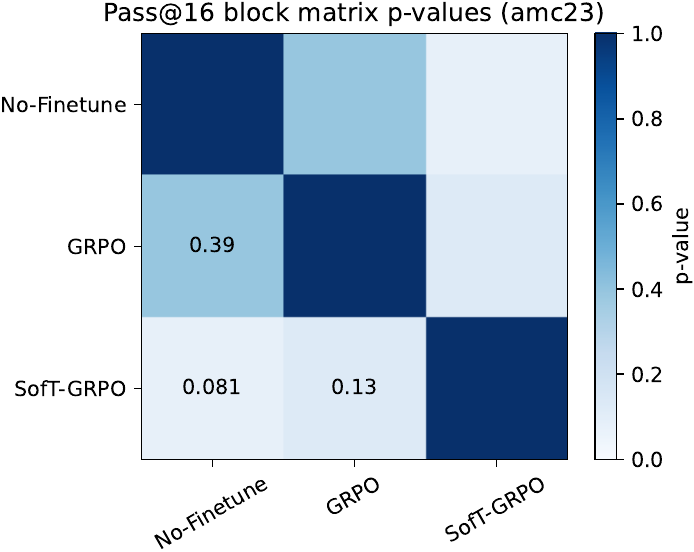}}\qquad\qquad
    \subfigure[P-scores of Pass@32 over SofT-GRPO, GRPO, and No-finetune on AMC23]{\includegraphics[width = 0.35\textwidth]{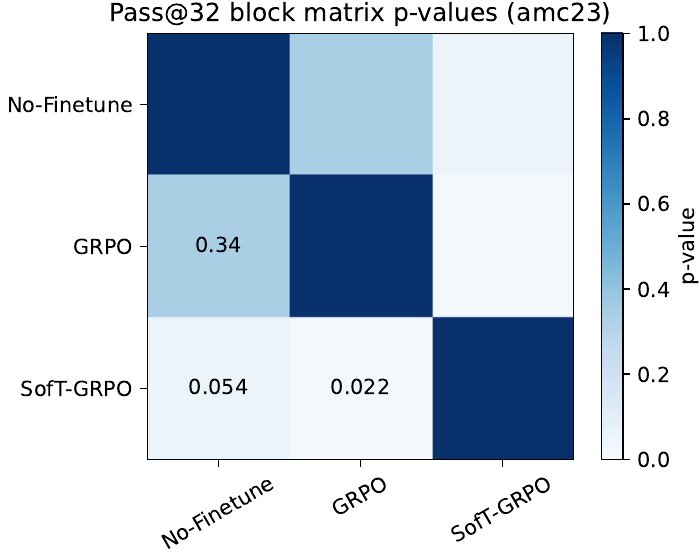}}\caption{Significance test on AMC23.}\label{fig:pamc}
\end{figure*}

Besides the results in Table \ref{maindata} and Figure \ref{fig:temperature}, in this section, we use results from 1024 runs to demonstrate the significance test for Pass@16 and Pass@32 (dividing 64 groups for Pass@16 from the 1024 runs and 32 groups of Pass@32). We calculated the t-test for the one-sided hypothesis test (See Figure \ref{fig:pass1024}, the average result of SofT-GRPO outperforms the GRPO and No-Finetune baselines), and taking 0.05 as the threshold, the results in Figure \ref{fig:paime} and \ref{fig:pamc} show that in Pass@16 with 64 runs, and Pass@32 with 32 runs:
\begin{itemize}
    \item \textbf{SofT-GRPO can significantly outperform GRPO on AIME2024 with Pass@16 (0.0067), AMC23 with Pass@32 (0.022)}, with quite small values (0.13) on both AIME2024 with Pass@32 and AMC23 with Pass@16.
    \item SofT-GRPO can get quite small p-values (0.081 and 0.054) on AMC23 with Pass@16 and Pass@32.
\end{itemize}


\newpage
\section{Discussion \& Analysis on Results}\label{Appendixe}

In this section, we will provide a detailed analysis of the improvement in Pass@K. \citet{yue2025does} first validates that RLVR techniques in discrete-token GRPO will not incentivize reasoning capacity in LLMs beyond the base model, demonstrating \textbf{equal or even inferior Pass@K values} when k improves. So, Recent studies \citep{walder2025pass, mahdavi2025beyond, peng2025simko} try to develop RL methods that optimize the Pass@K accuracy directly. Among them, \citet{walder2025pass,chow2024inference,mahdavi2025beyond,hu2026rewarding} focus on reallocating the reward within the group, encouraging rewards of negative discrete-token CoTs in groups with high Pass@K value. \citet{peng2025simko} try to improve the model's Pass@K performance by avoiding the over-concentration on the top-1 selection within the token set. 

In SimKO, at high-entropy (forking) tokens, positive (correct) samples are rewarded not only on the sampled token but spread across the top-K most probable tokens, promoting output diversity and better Pass@K. For negative (incorrect) samples, SimKO applies stronger penalties to the top-1 token and lighter penalties to others, which prevents the output probabilities from collapsing onto a single choice. This entropy-aware and asymmetric strategy encourages exploration, outperforming traditional RL methods that only focus on the sampled token.

In the proposed SofT-GRPO, we have the policy update in the soft-thinking part ($t$-th token) as:
\begin{align*}
\nabla_\theta L(\theta) 
&= \nabla_\theta \left( \sum_{i=1}^{|\mathcal{T}|} \left[ -g'_i + \log p_i - \exp(-g'_i + \log p_i) \right] - \left[ -\epsilon_i - \exp(-\epsilon_i) \right] \right) \\
&= \sum_{i=1}^{|\mathcal{T}|} \nabla\theta \left( \log p_i - \exp(-g'_i + \log p_i) \right) \\
&= \sum_{i=1}^{|\mathcal{T}|} \left( \nabla_\theta \log p_i - \exp(-g'_i + \log p_i) \nabla_\theta \log p_i \right) \\
&= \sum_{i=1}^{|\mathcal{T}|} \left( 1 - \exp(-g'_i + \log p_i) \right)\nabla_\theta \log p_i \\
&= \sum_{i=1}^{|\mathcal{T}|} \left( 1 - p_i \cdot \exp(-g'_i) \right)\nabla_\theta \log p_i.
\end{align*}
This means that instead of updating the policy only for the selected token, in SofT-GRPO, it inherently highlights: \textbf{(1)}. Higher importance component ($g'_i$) in the soft tokens, \textbf{(2)}. lower prob ($p_i$). For example, for positive instances, SofT-GRPO inherently improves the probability of the main component in each soft-tokens and the low-probability components. For Negative instances, SofT-GRPO inherently gives penalties to the probability of the main component in each soft-tokens and the low-probability components.

\begin{figure}[H]
    \centering
    \includegraphics[width=0.6\linewidth]{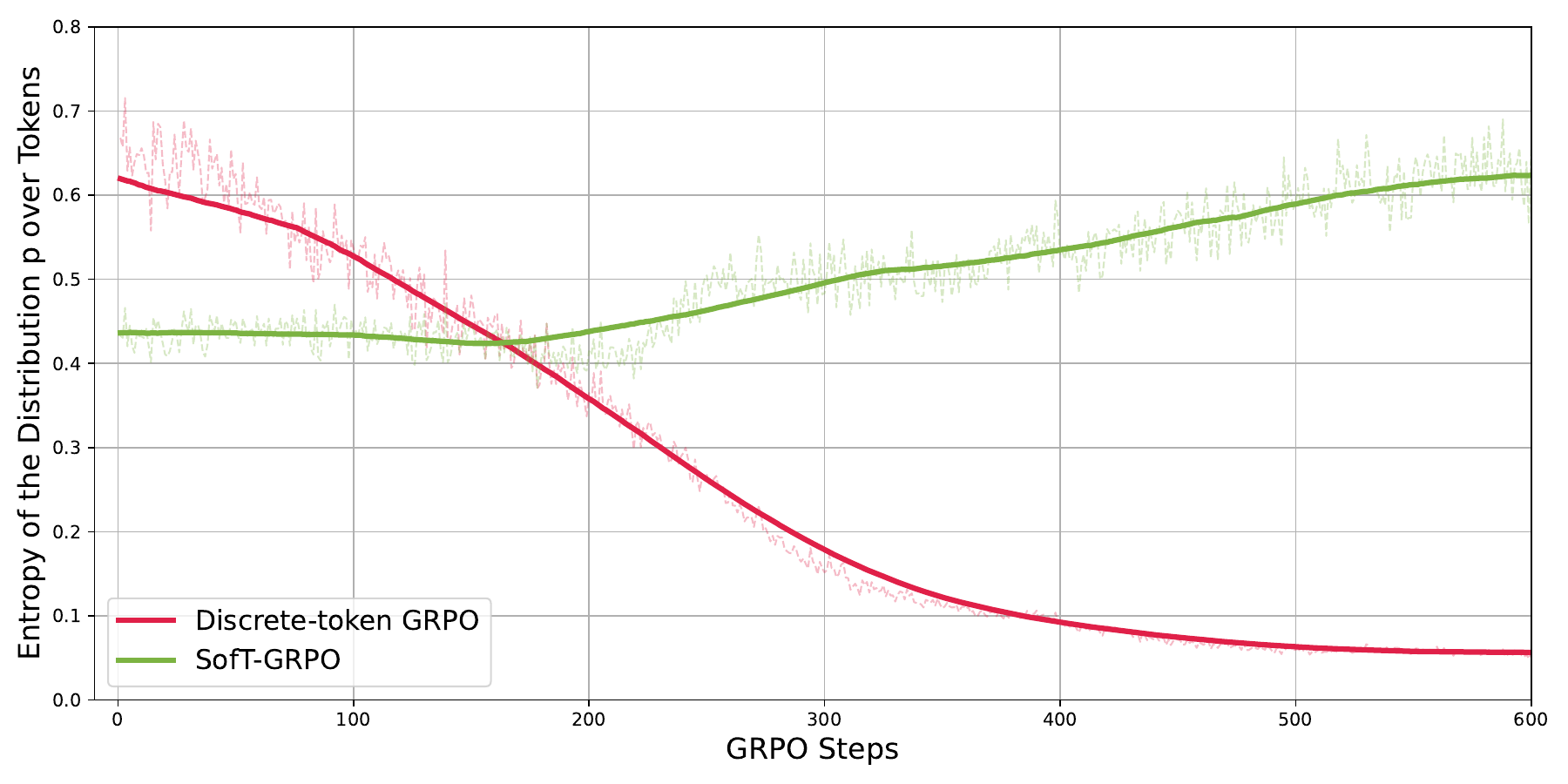}
    \caption{Entropy curve on \textit{\textbf{DeepSeek-R1-Distill-Qwen-7B}} during training.}
    \label{fig:llama-entropy}
\end{figure}

Figure \ref{fig:llama-entropy} verifies this conclusion: the entropy of token probability distributions will converge to nearly 0 with discrete-token GRPO, but keep stable and even improve with SofT-GRPO.

\newpage
\section{Baselines \& Datasets \& Licenses}
\label{Appendixf}

In our experiments, we evaluate and compare the following model baselines and datasets. For each, we detail the official website and usage license.

\subsection{Baselines}

We mainly include No-Finetune base LLMs, Discrete-Token GRPO, and the method in \citet{butt2025soft} (noted Soft Token) as baselines.

\paragraph{Base LLMs} This paper includes \textit{\textbf{DeepSeek-R1-Distill-Qwen-1.5B}}, \textit{\textbf{LLaMA-3.2-3B-Instruct}}, and \textit{\textbf{DeepSeek-R1-Distill-Qwen-7B}} as base LLMs.

\paragraph{Discrete-Token GRPO} We utilize the latest verl \citep{sheng2024hybridflow} \url{https://github.com/volcengine/verl/tree/main} and vLLM rollout to implement discrete-token GRPO with default parameters.

\paragraph{\citet{butt2025soft} Soft Token} Due to the requirement of passing d-dimensional inputs $\hat{\boldsymbol{s}}$ between the rollout workers and the verl policy optimization workers, implementing this algorithm requires a high amount of communication between the rollout workers and the policy update workers. So, we only report the results in \citep{butt2025soft} for comparison in Table \ref{ablation-main}.

\subsection{Datasets}

This paper covers five in-domain numerical reasoning datasets (i.e., AIME2024, AIME2025, AMC23, MATH-500, and GSM8K \citep{cobbe2021training}), one out-of-domain scientific reasoning dataset GPQA-Diamond \citep{rein2024gpqa}, and two out-of-domain code reasoning datasets (i.e., HumanEval and MBPP). These datasets are provided in \url{https://github.com/eric-ai-lab/Soft-Thinking}.

\subsection{Inference Framework}
The inference framework of our implementation is built on \textbf{SGLang}~\citep{zheng2024sglang}, maximizing efficiency via continuous batching, RadixAttention for KV cache reuse, and compressed finite state machines for faster structured output decoding. In the standard flow, the scheduler flattens discrete token inputs into continuous tensors for GPU execution; generated tokens are returned to update the Radix tree state for subsequent steps.

SGLang's optional overlap scheduler pipelines execution by issuing the next batch ahead via negative integer slot addresses. Compared to \citet{zhang2025soft}, \textbf{we enabled this} for Soft-Thinking by storing generated soft probabilities and indices in \textbf{VRAM-allocated future maps}. A custom kernel resolves the slot addresses to fetch these states just-in-time. This design \textbf{should be faster} by eliminating PCIe bottlenecks, allowing non-blocking scheduling with data kept entirely on the GPU.
\subsection{Licenses}

For Base LLM, Dataset, and frameworks, we list their Licenses in Table \ref{license}.

\begin{table}[H]
\centering
\setlength{\tabcolsep}{1mm}
\renewcommand\arraystretch{1.2}
\caption{A summary of licenses.}
\resizebox{\textwidth}{!}{
\begin{tabular}{llll}
\bottomrule[0.8mm]
Resources & Type    & License& URL    \\ \midrule[0.3mm]
DeepSeek-R1-Distill-Qwen-1.5B & Base LLM    &  MIT License &  \url{https://huggingface.co/deepseek-ai/DeepSeek-R1-Distill-Qwen-1.5B} \\
LLaMA-3.2-3B-Instruct & Base LLM    &  Llama 3.2 License &  \url{https://huggingface.co/meta-llama/Llama-3.2-3B-Instruct} \\
DeepSeek-R1-Distill-Qwen-7B & Base LLM    &   MIT License &  \url{https://huggingface.co/deepseek-ai/DeepSeek-R1-Distill-Qwen-7B} \\  \midrule[0.3mm]
verl & RL-framework    &  Apache-2.0 license &  \url{https://github.com/volcengine/verl} \\
verl-0.4.x & RL-framework    &  Apache-2.0 license &  \url{https://github.com/volcengine/verl/tree/v0.4.x} \\
SGLang for soft-thinking & Inference-framework    & MIT License     &\url{https://github.com/eric-ai-lab/Soft-Thinking}\\ \midrule[0.3mm]
AIME2024, AIME2025, GSM8K & Dataset    & MIT License    &\url{https://github.com/eric-ai-lab/Soft-Thinking} \\
AMC23, MATH-500 & Dataset    & Available Online    &\url{https://github.com/eric-ai-lab/Soft-Thinking} \\ 
HumanEval, GPQA-Diamond & Dataset    & MIT License    &\url{https://github.com/eric-ai-lab/Soft-Thinking} \\ 
MBPP & Dataset    & Apache-2.0 license   &\url{https://github.com/eric-ai-lab/Soft-Thinking} \\  \toprule[0.8mm]
\end{tabular}}
\label{license}
\end{table}

\newpage
\section{Visualization}
\label{Appendixg}
\begin{figure}[H]
    \qquad\qquad\includegraphics[width=0.85\linewidth]{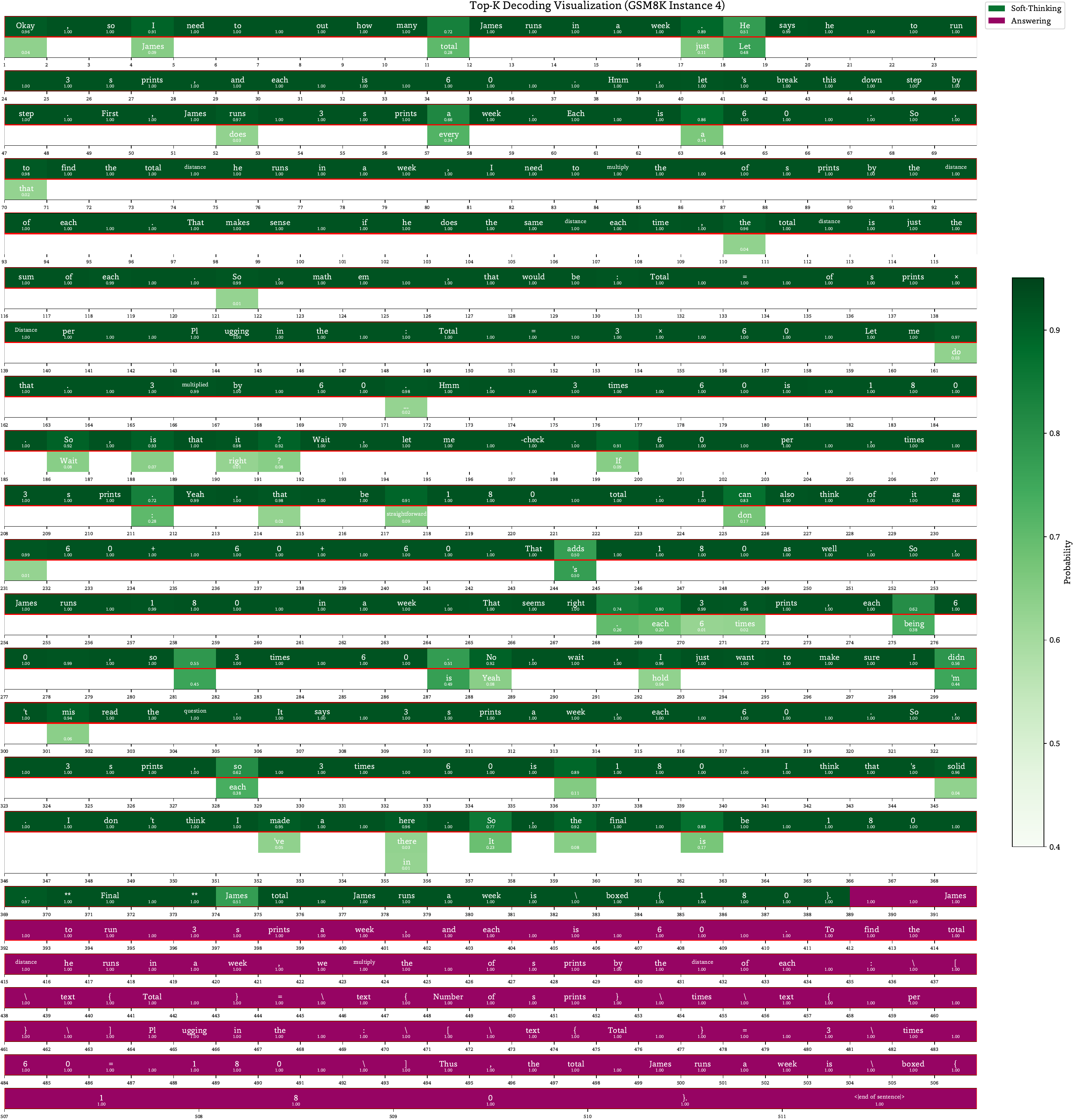}
    \caption{Visualization: \textbf{\textit{DeepSeek-R1-Distill-Qwen-1.5B}} + SofT-GRPO on the 4-th instance of GSM8K.}
    \label{fig:gsm8kexample}
\end{figure}

\begin{figure}
    \centering
    \includegraphics[width=\linewidth]{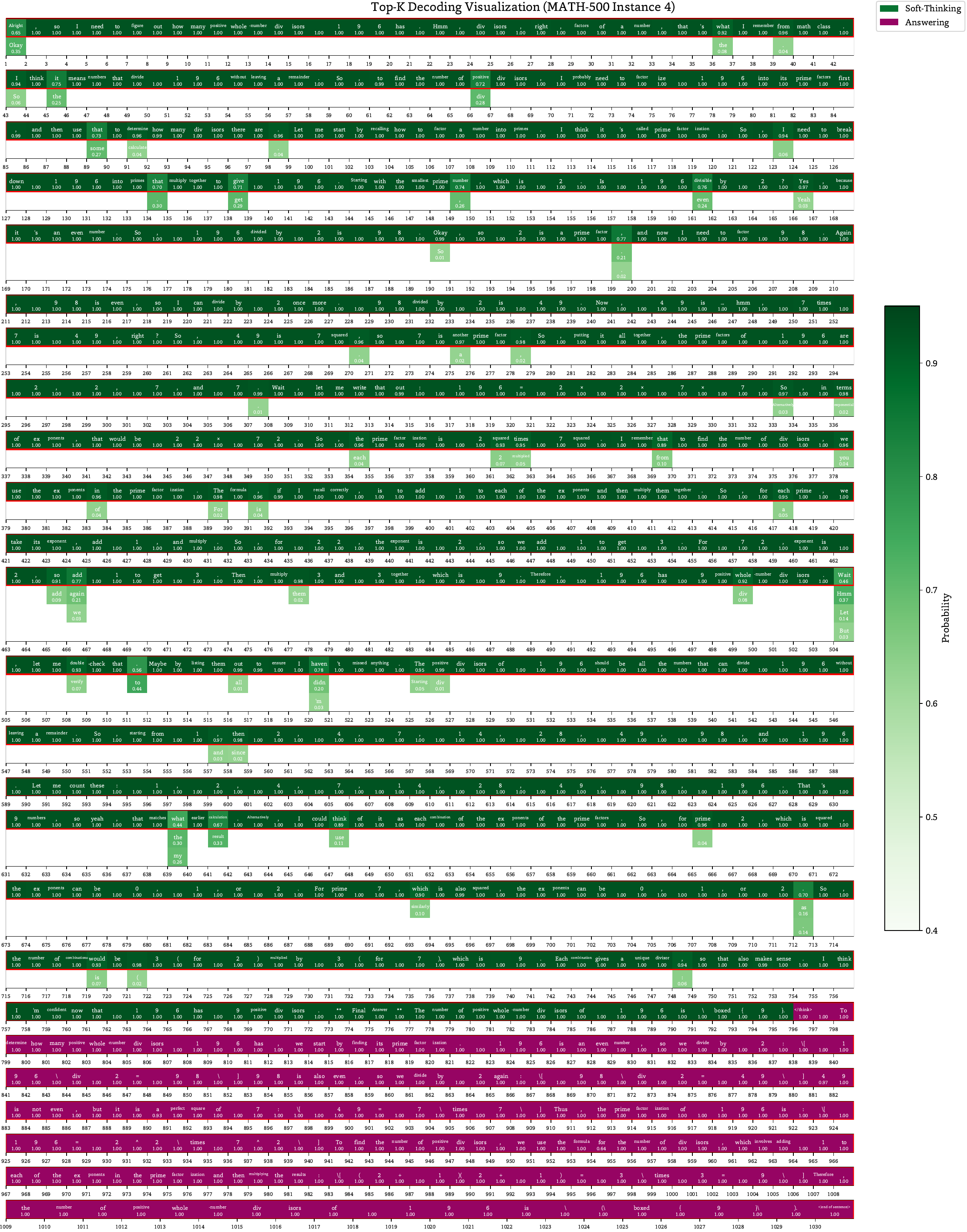}
    \caption{Visualization: \textbf{\textit{DeepSeek-R1-Distill-Qwen-1.5B}} + SofT-GRPO on the 4-th instance of MATH-500.}
    \label{fig:mathexample}
\end{figure}

In this section, in Figure \ref{fig:gsm8kexample} and Figure \ref{fig:mathexample}, we provide visualization examples for the outcome of SofT-GRPO on two examples (we show $y_i$ numbers that are larger than 0.01 in soft-tokens). Our observations are as follows:
\begin{itemize}
    \item Compared to Figure 4 in \citet{zhang2025soft}, SofT-GRPO preserves the multiply-path thinking pattern, thus providing better soft-thinking reasoning performances.
    \item As discussed in \citet{wu2025llms}, SofT-GRPO outcomes can observe possible latent search trees. See the 288th soft-token in Figure \ref{fig:gsm8kexample}, \textbf{it blends two discrete tokens with contrary meanings in its top-2} (No and Yeah, probably with totally different embeddings), representing a more abstract concept than any language token. Moreover, to check whether this pattern has already existed in No-Finetune LLM, we searched their output on GSM8K (a total of 1319 instances), where we could not find this probability distribution. \textbf{So, we can recognize that this probability distribution is highly possible to emerge after SofT-GRPO.}
    \item Similarly to Figure 4 in \citet{zhang2025soft}, SofT-GRPO can preserve interpretability; the contents in the red box (Okay, so I need to...) can be a good explanation for the path of soft-thinking reasoning.
\end{itemize}



\end{document}